\pdfoutput=1

\documentclass[5p,times]{elsarticle}

\usepackage{lineno,hyperref}
\modulolinenumbers[5]

\usepackage{amssymb}
\usepackage{amsmath}
\usepackage{tabularx}
\usepackage{multirow}
\usepackage{placeins}

\usepackage{framed}
\usepackage{fancyvrb}
\usepackage{algpseudocode}
\usepackage{algorithm}

\usepackage[utf8]{inputenc}
\usepackage[T1]{fontenc}
\usepackage{natbib}

\begin{document}

\begin{frontmatter}

\title{Introducing languid particle dynamics to a selection of PSO variants}


\author[mymainaddress]{Siniša Družeta}

\author[mymainaddress]{Stefan Ivić \corref{mycorrespondingauthor}}
\cortext[mycorrespondingauthor]{Corresponding author}
\ead{sivic@riteh.hr}

\author[mymainaddress]{Luka Grbčić}

\author[mymainaddress]{Ivana Lučin}

\address[mymainaddress]{University of Rijeka Faculty of Engineering, Vukovarska 58, Rijeka, Croatia}

\begin{abstract}
Previous research showed that conditioning a PSO agent's movement based on its personal fitness improvement enhances the standard PSO method. In this article, languid particle dynamics (LPD) technique is used on five adequate and widely used PSO variants. Five unmodified PSO variants were tested against their LPD-implemented counterparts on three search space dimensionalities (10, 20, and 50 dimensions) and 30 test functions of the CEC 2014 benchmark test. In the preliminary phase of the testing four of the five tested PSO variants showed improvement in accuracy. The worst and best-achieving variants from preliminary test went through detailed investigation on 220 and 770 combinations of method parameters, where both variants showed overall gains in accuracy when enhanced with LPD. Finally, the results obtained with best achieving PSO parameters were subject to statistical analysis which showed that the two variants give statistically significant improvements in accuracy for 13-50\% of the test functions.
\end{abstract}


\end{frontmatter}
\section{Introduction}
\label{sec:Introduction}
Particle Swarm Optimization (PSO) is an optimization method originally inspired by the movement of bird flocks and fish schools \cite{PSO1,PSO2}. The method tracks a group of agents (particles) moving through the search space, each agent adapting its movement on the basis of its own findings as well as the findings of other agents. To this day, a great number of  modifications and improvements have been proposed for PSO, as both elegance and capability of the method keep motivating researchers to further investigate its features and advance its performance.

In an attempt to enhance the effectiveness of PSO by enabling the particles with some awareness of their own improvement, we have previously proposed an enhancement of PSO method with Personal Fitness Improvement Dependent Inertia (PFIDI) which makes each particle's movement conditioned by its fitness improvement \cite{LPSO}. Although this being a fairly fundamental modification of the PSO particle movement logic, it neither undermines the elegance of PSO nor it betrays its bio-inspired origins. The proposed PFIDI method, as implemented in its most basic version called Languid PSO (LPSO), was shown to yield predominantly better accuracy than standard PSO, across a spectrum of various goal functions and method parameter configurations.

The results presented in \cite{LPSO}, however encouraging, possibly remain inconclusive due to the fact that PFIDI-enabled LPSO was tested only against standard PSO. Therefore, stronger evidence of the PFIDI/LPSO capabilities would be obtained if these effects were tested on several advanced PSO variants. In light of this, the performance of a selection of PSO variants enhanced with inertia handling technique used in LPSO is demonstrated in this article.

\section{Overview of PFIDI-related research}
\label{sec:Overview}
Although numerous advanced PSO variants exist which utilize some additional information on particle fitness and/or a kind of adaptive inertia technique, only those in which an individual particle's movement logic is directly influenced by its fitness and those which handle inertia in a particle-wise manner are understood as closely related to the technique explored in this article.

There are many PSO variants which track particles' fitness in order to use that information for local social attraction. One of these is the method proposed in \cite{BRAENDLER}, which combines fitness information with Euclidean distance so as to establish a neighborhood and find its best particle for \emph{lbest} PSO. However, for this only a particle's neighbors' fitness is considered, while the particle's own fitness is not used for its own self-assessment. A similar approach is the ``Fitness-Distance-Ratio'' method  \cite{FDRPSO} which finds a ``neighborhood best'' particle among a particle's spatially nearest neighbors, using a criterion of fitness-distance ratio. However, techniques like these institute a certain departure from the original bio-inspired idea of PSO of using simply behaving, dominantly autonomous particles, as well as often come with a significant computational cost.

As for the adaptation of inertia, various methods have been proposed, most of which understand inertia weight factor $w$ as a swarm-wise parameter, i.e. constant across the entire swarm \cite{IWSTRAT}. In the rare cases where particle-wise inertia weight adaptation is employed, inertia weight $w$ is usually calculated on the basis of relative difference between the fitness of an individual particle and some characteristic fitness of its neighborhood, such as the neighborhood best, average or worst fitness value \cite{CLERC,CPSO,RAGHAVENDRA,SCORE}. The adaptation is conducted so that more inertia is provided to particles with relatively bad positions and less inertia to particles with relatively good positions. A different approach, yielding similar effect, is proposed in \cite{DONG}, where the calculation of $w$ is based on a fitness-based particle ranking. The principal idea behind all these inertia weight adaptation methods is that speeding up low-performing particles and slowing down high-performing particles should improve the search effectiveness of the entire swarm.

A different idea, partly related to the fitness improvement based inertia adaptation, is implemented in the Inertia-Adaptive PSO  \cite{IAPSO}, where inertia factor $w$ is adapted separately for each particle, according to the distance between the particle in question and its neighborhood's historically best position. This technique ensures that the ``diverging'' particles are even more strongly attracted towards neighborhood best and personal best positions. A somewhat similar method is used also in \cite{FENG}. Here a particle's inertia weight factor value depends on its velocity vector and resultant attraction vector towards both personal best and neighborhood best positions, with both directions and magnitudes of the vectors taken into account. Effectively, inertia weight factor of each particle depends on its velocity and position, with respect to personal best and neighborhood best positions.

The rate of personal fitness improvement has also been proposed as a criterion for particle-wise adaptive inertia \cite{ZHANG,YANG}. In this technique, personal ``evolution speed'' is assessed, together with fitness-based ``swarm aggregation degree'', so as to be used to dynamically calculate inertia weight factor value.

Another approach, rather closely related to the idea of PFIDI, is the inertia handling for multi-swarm PSO proposed in \cite{DYNAMIC}. Here the inertia weight values are constant across each swarm, but differrent between the swarms as they are dynamically updated, depending on the number of particles in the swarm with improving fitness. In effect, when the particles mostly improve, the inertia of the entire swarm is amplified and vice versa, making this a de facto swarm-wise version of fitness improvement dependent inertia.

The inertia weight adaptation in \cite{LAIW} appears to be particularly interesting for the research of PFIDI. This technique employs fitness based inertia control both on a global level and on a local level (i.e. per-particle). In the local version (called Locally Adaptive Inertia Weight) the inertia weight of each particle is obtained as a difference between the fitness of its current personal best position and the fitness of the personal best position of the previous iteration. Although the straightforwardness of the technique makes it rather attractive, it also raises questions of sensitivity to fitness value magnitude, which is certainly not desirable, at least for general-purpose optimization methods.

\section{Languid particle dynamics in standard PSO}
\label{sec:Languidism}

\subsection{Standard PSO}
\label{sec:SPSO}

In standard PSO the particles move with a certain amount of inertia through the search space, while being attracted to the best position that they individually have found, and to the best position found by any particle of their neighborhood.

For each individual particle of the PSO swarm, we keep track of its position in the $D$-dimensional search space $\mathbf{x} = (x_1, x_2, x_3, ... x_D)$, its historically best position $\mathbf{p}$, its current velocity $\mathbf{v}$ and historically best position of its neighboring particles $\mathbf{g}$. After random initialization of positions $\mathbf{x}$ and velocities $\mathbf{v}$, a $k$-th particle moves by updating its velocity and position at iteration $t$ \cite{PSO98}:
\begin{equation}
\begin{aligned}
\label{eq:PSOv}
\mathbf{v}_k^{(t)} ={} & \mathbf{v_I}_k^{(t)} + c_1\cdot \mathbf{r}_1\circ(\mathbf{p}_k^{(t-1)}-\mathbf{x}_k^{(t-1)})\\  
& + c_2\cdot \mathbf{r}_2\circ(\mathbf{g}_k^{(t-1)}-\mathbf{x}_k^{(t-1)})~,
\end{aligned}
\end{equation}
\begin{equation}
\label{eq:PSOx}
\mathbf{x}_k^{(t)} = \mathbf{x}_k^{(t-1)} + \mathbf{v}_k^{(t)}~,
\end{equation}
where $\mathbf{v_I}$ is inertial velocity, $c_1$ and $c_2$ are cognitive and social PSO coefficients, respectively, while $\mathbf{r}_1$ and $\mathbf{r}_2$ are  $D$-dimensional vectors of random numbers in the range $[0,1]$. Note that vector multiplication in \eqref{eq:PSOv} is a Hadamard product.

In standard PSO, inertial velocity $\mathbf{v_I}$ is defined as the particle's velocity of the previous iteration, scaled with a weight factor:
\begin{equation}
\label{eq:vi}
\mathbf{v_I}_k^{(t)} = w^{(t)} \cdot \mathbf{v}_k^{(t-1)}~,
\end{equation}
with the purpose of the inertia weight $w$ being to serve as a control mechanism for swarm convergence.
Since different optimization problems require different convergence dynamics, many various methods have been proposed for dynamically changing or adapting inertia weight \cite{IWSTRAT}. Still, inertia weight is often used as a constant and then it is generally recommended to use $w = 0.7 \pm 0.05$ \cite{PHDTHESIS,STANDARDPSO}.

Coefficients $c_1$ and $c_2$ are traditionally used as $c_1 = c_2 = 2.0$, but better understanding of their influence encourages the use of lower values and problem-specific calibration \cite{PHDTHESIS,STANDARDPSO}, as well as changing them over iterations \cite{TVAC}.

The version of PSO with each particle being informed by the entire swarm is called `\emph{gbest} PSO', while the version with each particle communicating only with a subset of the swarm is called `\emph{lbest} PSO' \cite{PSO2}. In other words, a swarm may be entirely connected in a single neighborhood or divided in many smaller neighborhoods, with neighborhood topologies being purely index-based, i.e. not related to search space locality. Many different options have been proposed for the neighborhood topology of the \emph{lbest} PSO and no specific topology has been universally adopted as most beneficial in terms of overall PSO performance. Standard PSO implementations mostly imply the use of simple circular (`ring') topology \cite{STANDARDPSO} or random topology \cite{PSOCODE}.

Note that, since standard PSO particles do not track their fitness progress, they have no information on the fitness change along their path. In the course of PSO research, a number of techniques for using fitness improvement information for improving the efficiency of the swarming process have been developed, albeit only a few of those consider particle's own fitness improvement in each particle's movement logic. On the other hand, although some authors have proposed particle-wise inertia control, in standard PSO as well as in most other PSO variants inertia is considered to be global (i.e. constant across the entire swarm). Based on utilizing particle-wise inertia control for resolving personal fitness improvement dependent particle movement, the PFIDI approach and the LPSO method \cite{LPSO} were proposed.

\subsection{Languid particle dynamics}
\label{sec:Languid}

In an attempt to enhance PSO performance a PFIDI technique is proposed in which each particle tracks its fitness evolution and then uses this information in its movement decision-making process. One possible technique for this is using fitness improvement as a prerequisite for particle inertia. As probably the simplest method for this, a basic switch-like condition on inertia term \ref{eq:vi} of each individual particle \cite{LPSO} is used:
\begin{equation}
\label{eq:viLPSO}
	\mathbf{v_I}_k^{(t)} =
	\begin{cases}
		(w^{(t)}+0.05) \cdot \mathbf{v}_k^{(t-1)} & \text{when 		$f(\mathbf{x}_k^{(t-1)}) < f(\mathbf{x}_k^{(t-2)})$} \\
		\mathbf{0} &\text{otherwise}
	\end{cases}~,
\end{equation}
where $f$ is the fitness function. This means that the $k$-th particle has inertia only as long as it keeps advancing in a direction of better fitness (the formulation \eqref{eq:viLPSO} assumes a minimization problem).

Behaving in this manner, a particle disregards its previous direction if it failed to take it to a better location. Since such particle behavior implies a certain lack of enthusiasm, the adjective `languid' was adopted as a designator of this type of particle movement dynamics. Respectively, standard PSO enabled with languid particle dynamics (LPD) is named Languid PSO (LPSO).

Note that \eqref{eq:viLPSO} employs a correction of +0.05 for inertia weight when inertia is not disabled. This represents a modification of the original inertia handling technique of Languid PSO  \cite{LPSO}, which used no such correction. Since the originally proposed inertia switching technique reduces the overall velocity of the swarm, it would be reasonable to expect that the method would benefit from compensation in the form of increased inertia weight. This was also indicated by the results of the benchmark testing given in previous research \cite{LPSO}, thus the aforementioned amplification coefficient was introduced in order to address this issue.

The pseudo-code of standard PSO enhanced with languid particle dynamics is given in Algorithm 1.

\begin{algorithm}[H]
\caption{Standard PSO with languid particle dynamics}
\begin{algorithmic}[1]
\For{particle $k = 1$ \textbf{to} $n$}
\State number of evaluations $eval=0$
\State iteration $t=0$
\State initialize particle position $\mathbf{x}_k^{(t)}$ and velocity $\mathbf{v}_k^{(t)}$
\State initialize neighborhood topology
\State evaluate fitness $f(\mathbf{x}_k^{(t)})$, $eval = eval + 1$
\State find personal and neighborhood best: $\mathbf{p}_k^{(t)}$, $\mathbf{g}_k^{(t)}$
\EndFor
\While{$eval < eval_{max}$}
\State iteration $t=t+1$
\For{particle $k = 1$ \textbf{to} $n$}
\If{$f(\mathbf{x}_k^{(t-1)}) < f(\mathbf{x}_k^{(t-2)})$ \textbf{or} $t < 2$}\Comment{Eq. \eqref{eq:viLPSO}}
\State $\mathbf{v_I}_k^{(t)} = (w^{(t)}+0.05) \cdot \mathbf{v}_k^{(t-1)}$
\Else
\State $\mathbf{v_I}_k^{(t)} = \mathbf{0}$
\EndIf
\State calculate new velocity $\mathbf{v}_k^{(t)}$ \Comment{Eq. \eqref{eq:PSOv}}
\State calculate new position $\mathbf{x}_k^{(t)}$ \Comment{Eq. \eqref{eq:PSOx}}
\State evaluate fitness $f(\mathbf{x}_k^{(t)})$, $eval = eval + 1$
\State find personal and neighborhood best: $\mathbf{p}_k^{(t)}$, $\mathbf{g}_k^{(t)}$
\EndFor
\If{$f^{(t)}_{best} > f^{(t+1)}_{best}$ and $lbest$}
\State reinitialize neighborhood topology
\EndIf
\EndWhile\label{euclidendwhile}
\end{algorithmic}
\end{algorithm}

\section{Selected PSO variants}
\label{sec:SelectedPSO}

In order to thoroughly test the effects of LPD (i.e. LPSO inertia switching technique), a selection of PSO variants was used. For each variant a sub-variant enabled with LPD was implemented and then had its performance compared with the pure variant. The PSO variants used are:

\begin{itemize}\itemsep0em
	\item Linearly Decreasing Inertia Weight PSO (LDIW-PSO)
	\item Time Varying Acceleration Coefficients PSO (TVAC-PSO)
	\item Chaotic PSO (C-PSO)
	\item Dynamic Multiswarm PSO (DMS-PSO)
	\item Comprehensive Learning PSO (CL-PSO).
\end{itemize}

The above variants not only are some of the most popular in PSO research and applications, but also employ various interesting PSO advancement techniques.
As such, these variants make a good assortment of PSO modifications for testing the compatibility of LPD with other PSO enhancement techniques.

\subsection{Linearly Decreasing Inertia Weight PSO (LDIW-PSO)}
\label{sec:LDIWPSO}

Linearly Decreasing Inertia Weight PSO (LDIW-PSO) was introduced in \cite{LDIW} as an improvement over standard PSO which uses constant inertia weight factor. In LDIW-PSO intertia weight factor $w$ is linearly decreasing with every iteration from its maximum value to its minimum value, which are both defined at the begining of the optimization. This way the method gradually transitions from global to local search.

Due to its beneficient effects on PSO accuracy and convergence, LDIW technique has become widely used, both in PSO research and application.

Particle positions and velocities at iteration $t$ are updated as given in \eqref{eq:PSOv} and \eqref{eq:PSOx}, with inertia weight factor used in \eqref{eq:vi} being updated at every iteration as follows:
\begin{equation}
\label{eq:wLDIW}
w^{(t)} = w_{min}+(w_{max}-w_{min}) \frac{t}{t_{max}}
\end{equation}
where $w_{min}$ is the minimum value of interia weight factor, $w_{max}$ is the maximum value of intertia weight factor, $t$ is the current iteration for which $w$ is calculated, and $t_{max}$ is the maximum number of allowed iterations (corresponding to maximum allowed function evaluations $eval_{max}$). 

In this research $w_{min}=0.4$ and $w_{max}=w_0+0.2$ were used, with $w_0$ values for calculating $w_{max}$ being taken from a range of inertia weight factor values (Tables \ref{tab:MethodParameters} and \ref{tab:ParameterRanges}). Varying of $w_{max}$ in this manner allows for fine-tuning of the upper limit of the linearly decreasing inertia.

\subsection{Time Varying Acceleration Coefficients PSO (TVAC-PSO)}
\label{sec:TVACPSO}

Time varying acceleration coefficients PSO (TVAC-PSO) \cite{TVAC} is a PSO variant which uses linearly changing PSO coefficients $c_1$ and $c_2$, contrary to the standard PSO where coefficients $c_1$ and $c_2$ are used as constant values. Coefficients $c_1$ and $c_2$ calculated by:
\begin{equation}
\label{eq:TVACc1}
c_1^{(t)} = c_{1_i} + (c_{1_f}-c_{1_i}) \frac{t}{t_{max}}
\end{equation}
\begin{equation}
\label{eq:TVACc2}
c_2^{(t)}  = c_{2_i} + (c_{2_f}-c_{2_i}) \frac{t}{t_{max}}
\end{equation}
where $c_{1_i}$, $c_{1_f}$, $c_{2_i}$ and $c_{2_f}$ represent initial and final values of coefficients $c_1$ and $c_2$ which are linearly increasing/decreasing over iterations of the PSO swarming process. 

In the TVAC-PSO implementation used for this research $c_1$ is decreasing from $c_{1_i}=2.5$ to $c_{1_f}=0.5$ while $c_2$ is increasing from $c_{2_i}=0.5$ to $c_{2_f}=2.5$, as recommended by the authors of the method. This approach encourages a wide global search in the early part of optimization and a detailed local search in the later phase of the optimization process, facilitating an efficient convergence to a global optimum. 

TVAC-PSO also features linearly decrasing inertia weight factor, as given in \eqref{eq:wLDIW}, coupled with standard PSO velocity \eqref{eq:PSOv} and position updating \eqref{eq:PSOx}.

The efficiency of TVAC-PSO, coupled with the reduced number of method parameters it provides, makes it a very attractive PSO variant and as such it was included in this research.

\subsection{Chaotic PSO (C-PSO)}
\label{sec:CPSO}

Chaotic PSO (C-PSO) was introduced in \cite{CPSO} and it has become one of the most popular PSO variants, used on a variety of optimization problems. 

In contrast to the standard PSO, C-PSO employs an additional secondary search called chaotic local search (CLS). So as to further improve the fitness of the best particle, CLS is used for exploring its vicinity after each PSO iteration.

In the C-PSO implementation used in this study, linearly decreasing inertia weight factor mechanism \eqref{eq:wLDIW} was used, together with standard PSO velocity \eqref{eq:PSOv} and position updating \eqref{eq:PSOx}.

After each PSO iteration, the C-PSO algorithm reserves the top $n/5$ of all particles sorted by fitness, where $n$ is the total number of particles. CLS is performed on the best particle of the entire swarm ($\mathbf{g}$), while the other $4n/5$ particles are randomly generated (reinitialized).

In order to use the chaotic logistic map needed for CLS, optimization variables must be mapped to chaotic variables since they must be in a predefined range. Hence for each $x \in \mathbf{g}$ the optimization variables $x \in (x_{min}, x_{max})$ (with $x_{min}$ and $x_{max}$ being their lower and upper bounds) are converted to chaotic variables $\xi^{(0)} \in \langle0,1\rangle$ by:

\begin{equation}
\label{eq:real2chaotic}
\xi^{(0)} = \frac{x-x_{min}}{x_{max} - x_{min}} .
\end{equation}

After this conversion, CLS is performed by iterating on the logistic map defined as:
\begin{equation}
\label{eq:chaosmap}
\xi^{(i+1)} = \mu \cdot \xi^{(i)}(1-\xi^{(i)})
\end{equation}
where $i \in \{0, 1, ..., 9\}$ is the iteration number of the CLS. Starting value $\xi^{(0)} \notin \{0.25, 0.5, 0.75\}$ and the choice of $\mu = 4$ ensure chaotic behavior of the equation. After each CLS iteration, a newly discovered position needs to be mapped back to the original search space:
\begin{equation}
\label{eq:chaotic2real}
x = x_{min} +  \xi^{(i+1)} (x_{max}-x_{min})
\end{equation}
and the new solution $\mathbf{g}^{CLS} = (x_1, x_2, x_3, ... x_D)$ is evaluated. If $f(\mathbf{g}^{CLS})<f(\mathbf{g})$, the $\mathbf{g}^{CLS}$ position is adopted as new $\mathbf{g}$, otherwise the CLS loop \eqref{eq:chaosmap} is continued until the maximum number of CLS iterations has been reached ($i<10$).

Implemented in this manner, chaotic local search serves as an effective enhancement of the PSO method.

\subsection{Dynamic Multiswarm PSO (DMS-PSO)}
\label{sec:DMSPSO}

Dynamic multiswarm PSO (DMS-PSO) \cite{DMSPSO} represents a certain radicalization of the \emph{lbest} PSO idea. In DMS-PSO the swarm is divided into several subswarms, i.e. neighborhoods of fully connected  particles with no connections between the neighborhoods. During the optimization process the subswarms are dynamically reorganized so as to avoid premature convergence.

In the DMS-PSO implementation used for this research, each subswarm consists of $\max(5, n/10)$ randomly chosen particles, so as to avoid miniscule subswarms, while allowing for linear subswarm division for larger swarm size $n$. Furthermore, the used DMS-PSO implementation does not use fixed or random regrouping periods as originally proposed in \cite{DMSPSO}, but regroups the subswarms at every iteration at which the entire swarm failed to improve its best position.

DMS-PSO employs a very logical technique for handling PSO topology, which in turn shows good performance, while still allowing for possible improvements via advanced inertia handling such as PFIDI.

\subsection{Comprehensive Learning PSO (CL-PSO)}
\label{sec:CLPSO}

CL-PSO \cite{CLPSO} improves the efficiency of PSO by removing the need for using globally best position $\mathbf{g}$ in the velocity update equation \eqref{eq:PSOv}. This is acheived by performing the velocity update as:
\begin{equation}
\label{eq:CLPSOv}
\mathbf{v}_k^{(t)} = \mathbf{v_I}_k^{(t)} + c \cdot \mathbf{r}_k \circ (\mathbf{q}_k^{(t-1)} - \mathbf{x}_k^{(t-1)})
\end{equation}
where $c$ is the method coefficient, and $\mathbf{q}$ denotes the $k$-th particle's ``exemplar'' position. 

The ``exemplar'' positions $\mathbf{q}_k = (q_{k,1}, q_{k,2}, ..., q_{k,D})$ are obtained as an assortment of own and other particles' personal best values $\mathbf{p}$, as:
\begin{equation}
\label{eq:CLPSOq}
q_{k,d} = 
\begin{cases}
p_{l,d} & \text{for $r<P_k$} \\
p_{k,d} &\text{otherwise}
\end{cases}~,\;\text{for } d=1,2,...,D.
\end{equation}
Here $r$ is a randomly generated number $r \in [0, 1]$ and $P_k$ is the learning probability for the $k$-th particle, calculated by use of the following formula:
\begin{equation}
\label{eq:CLPSOprob}
P_k = 0.05 + 0.45 \frac{e^{\frac{10(k-1)}{n-1}} - 1}{e^{10} - 1}~.
\end{equation}
Each ``exemplar'' particle, denoted by $l$ in \eqref{eq:CLPSOq}, is selected by tournament selection with tournament size being equal to two.

If it happens that all ``exemplars'' of a particle are its own (i.e. $\mathbf{q}_k = \mathbf{p}_k$), a variable $d$ is chosen randomly from $\{1, 2, ..., D\}$ and the $q_{k,d} =p_{l,d}$ is applied with $l$ determined by size-2 tournament selection.

Regarding inertia handling, CL-PSO also uses linearly decreasing inertia weight, as per \eqref{eq:wLDIW}.

\section{Benchmark testing}
\label{sec:Benchmark}

For each of the selected PSO variants the performance of a pure variant was compared with the performance of the languid version of the variant. This was executed by use of an extensive benchmark testing program, based on the CEC 2014 test. CEC 2014 test was designed for benchmarking of real-parameter single objective optimization algorithms and comprises 30 test functions, most of which have randomly shifted global optima, while all are randomly rotated (see \cite{CEC2014} for details). The test consists of:

\begin{itemize}\itemsep0em
	\item unimodal functions (F1, F2, and F3)
	\item shifted multimodal functions (F4, F5, ..., F16)
	\item hybrid functions based on unimodal functions and shifted multimodal functions (F17, F18, ..., F22)
	\item composition functions based on unimodal functions, shifted multimodal functions and hybrid functions (F23, F24, ..., F30).
\end{itemize}

There are sufficient arguments for using CEC 2014 test for the testing of the effects of LPD on PSO. Firstly, CEC 2014 test was not designed specifically for the purpose of testing the effects of LPD, PFIDI, nor PSO in general, hence no inherent bias towards the tested methods and techniques is to be expected. Secondly, the extent of CEC 2014 test should provide strong evidence on the existence of any benefits of LPD for PSO accuracy.

The experimental tests were conducted on three search space dimensionalities, namely $D \in \{ 10, 20, 50 \}$.

In order to minimize any risks of software bugs, the selected PSO variants were implemented as modifications of the `standard PSO' code \cite{PSOCODE,SPSO}, which was modified only as much as was necessary for implementing the used variants.

In the variants which allow for a \emph{lbest} version, random topology was applied and neighborhood radius $K = 3$ was used (meaning that each particle has a neighborhood of $K$ other randomly chosen particles). Furthermore, in all variants which allowed for this, best-of-swarm fitness value was checked at every iteration and neighborhood randomization was triggered whenever best-of-swarm fitness value failes to improve (as implemented in \cite{PSOCODE}).

\subsection{Preliminary testing of languid particle dynamics effects on selected PSO variants}
\label{sec:PreliminaryTesting}

As a first phase of the benchmark testing, the overall effects of the implementation of LPD on the used PSO variants were assessed.

Considering that swarm size $n$, PSO coefficients $w_0$ and $c$ (where $c = c_1 = c_2$), as well as the choice between \emph{gbest} and \emph{lbest} PSO realistically need to be treated as problem-specific, it was decided to use the best performing method parameters for standard PSO on CEC 2014 test functions, as they were determined in previous research \cite{LPSO}. Table \ref{tab:MethodParameters} gives these parameters for the three search space dimensionalities. It should be noted that these parameters were used as the tested PSO variants allowed, i.e. $w_0$ values were used for linearly decreasing inertia weight as explained in subsection \ref{sec:LDIWPSO}, listed $c$ values were not used for TVAC-PSO and  \emph{gbest}/\emph{lbest} versions were not used for CL-PSO and DMS-PSO. 

Although the parameters given in Table \ref{tab:MethodParameters} are optimal for standard PSO and not necessarily also for the PSO variants used in this article, it is reasonable to believe that they are at least near-optimal for some of the used PSO variants. In light of this, it was only sensible to use them in this research, while admitting that the usage of best performing parameters is not fundamental for testing the effects of LPD.

\begin{table*}[!]\footnotesize
\caption{Best performing PSO parameters \cite{LPSO}}
\label{tab:MethodParameters}
\begin{tabular*}{\textwidth}{l@{\extracolsep{\fill}}rrrrrrrrrrrr}
\hline
\multicolumn{1}{c}{\multirow{2}{*}{Function}}  & \multicolumn{4}{c}{$D = 10$} & \multicolumn{4}{c}{$D = 20$}  & \multicolumn{4}{c}{$D = 50$} \\
\multicolumn{1}{c}{} & \multicolumn{1}{c}{$n$} & \multicolumn{1}{c}{$w_0$} & \multicolumn{1}{c}{$c$} & \multicolumn{1}{c}{version} & \multicolumn{1}{c}{$n$} & \multicolumn{1}{c}{$w_0$} & \multicolumn{1}{c}{$c$} & \multicolumn{1}{c}{version} & \multicolumn{1}{c}{$n$} & \multicolumn{1}{c}{$w_0$} & \multicolumn{1}{c}{$c$} & \multicolumn{1}{c}{version} \\
\hline
\multicolumn{1}{c}{F1} &  30 & 0.65 & 1.00 & \emph{gbest} & 40 & 0.80 & 0.75 & \emph{gbest} &  30 & 0.70 & 1.00 & \emph{lbest}\\ 
\multicolumn{1}{c}{F2} & 100 & 0.75 & 0.50 & \emph{lbest} & 20 & 0.50 & 1.25 & \emph{lbest} & 100 & 0.80 & 0.75 & \emph{lbest}\\ 
\multicolumn{1}{c}{F3} &  80 & 0.60 & 1.00 & \emph{gbest} & 60 & 0.65 & 1.00 & \emph{gbest} &  30 & 0.65 & 0.75 & \emph{lbest}\\ 
\multicolumn{1}{c}{F4} &  10 & 0.70 & 1.25 & \emph{gbest} & 30 & 0.85 & 0.75 & \emph{gbest}  & 120 & 0.55 & 1.25 & \emph{gbest}\\ 
\multicolumn{1}{c}{F5} &  25 & 0.50 & 1.00 & \emph{lbest} & 30 & 0.55 & 1.00 & \emph{gbest}  &  60 & 0.50 & 1.25 & \emph{gbest}\\ 
\multicolumn{1}{c}{F6} &  60 & 0.60 & 1.00 & \emph{lbest} & 120 & 0.65 & 1.00 & \emph{lbest} & 170 & 0.55 & 1.25 & \emph{lbest}\\ 
\multicolumn{1}{c}{F7} &  80 & 0.55 & 0.75 & \emph{lbest} & 120 & 0.65 & 1.00 & \emph{lbest} &  80 & 0.60 & 1.25 & \emph{lbest}\\ 
\multicolumn{1}{c}{F8} &  40 & 0.55 & 1.25 & \emph{lbest} & 60 & 0.55 & 1.25 & \emph{lbest} & 120 & 0.50 & 1.25 & \emph{lbest}\\ 
\multicolumn{1}{c}{F9} &  40 & 0.60 & 1.00 & \emph{lbest} & 60 & 0.70 & 0.75 & \emph{lbest} &  60 & 0.50 & 1.25 & \emph{lbest}\\ 
\multicolumn{1}{c}{F10} &  20 & 0.55 & 1.50 & \emph{lbest} & 120 & 0.55 & 1.75 & \emph{gbest} & 200 & 0.60 & 1.75 & \emph{gbest} \\ 
\multicolumn{1}{c}{F11} &  30 & 0.65 & 0.75 & \emph{lbest} & 120 & 0.60 & 1.75 & \emph{gbest} & 200 & 0.75 & 1.25 & \emph{gbest}\\ 
\multicolumn{1}{c}{F12} &  25 & 0.60 & 1.50 & \emph{gbest} & 80 & 0.50 & 1.75 & \emph{gbest} &  60 & 0.50 & 1.75 & \emph{gbest}\\ 
\multicolumn{1}{c}{F13} & 100 & 0.50 & 1.00 & \emph{lbest} & 120 & 0.55 & 1.00 & \emph{lbest} & 200 & 0.50 & 1.25 & \emph{lbest}\\ 
\multicolumn{1}{c}{F14} & 100 & 0.50 & 1.50 & \emph{lbest} & 140 & 0.60 & 1.00 & \emph{lbest} & 200 & 0.55 & 1.25 & \emph{lbest}\\ 
\multicolumn{1}{c}{F15} & 100 & 0.60 & 1.25 & \emph{gbest} & 100 & 0.65 & 1.50 & \emph{gbest} & 200 & 0.60 & 1.50 & \emph{gbest}\\ 
\multicolumn{1}{c}{F16} &  25 & 0.85 & 0.50 & \emph{lbest} & 25 & 0.65 & 1.75 & \emph{gbest}  & 200 & 0.85 & 0.50 & \emph{gbest}\\ 
\multicolumn{1}{c}{F17} &  30 & 0.80 & 0.50 & \emph{gbest} &  50 & 0.85 & 0.50 & \emph{gbest} & 40 & 0.80 & 0.50 & \emph{lbest}   \\ 
\multicolumn{1}{c}{F18} & 100 & 0.50 & 1.25 & \emph{lbest} & 40 & 0.70 & 0.75 & \emph{lbest} & 200 & 0.75 & 0.75 & \emph{lbest}\\ 
\multicolumn{1}{c}{F19} &  60 & 0.55 & 1.00 & \emph{lbest} & 100 & 0.75 & 0.75 & \emph{lbest} & 60 & 0.70 & 1.25 & \emph{gbest}\\ 
\multicolumn{1}{c}{F20} & 100 & 0.55 & 1.25 & \emph{lbest} & 60 & 0.65 & 1.00 & \emph{lbest} &  30 & 0.80 & 0.75 & \emph{lbest}\\ 
\multicolumn{1}{c}{F21} &  30 & 0.70 & 1.25 & \emph{lbest} & 120 & 0.65 & 1.00 & \emph{lbest} &  30 & 0.80 & 0.50 & \emph{lbest}\\ 
\multicolumn{1}{c}{F22} &  30 & 0.55 & 1.50 & \emph{lbest} & 60 & 0.60 & 1.00 & \emph{lbest} &  80 & 0.70 & 0.75 & \emph{lbest}\\ 
\multicolumn{1}{c}{F23} &  40 & 0.50 & 1.00 & \emph{gbest} & 40 & 0.50 & 1.25 & \emph{lbest} & 100 & 0.50 & 1.00 & \emph{lbest}\\ 
\multicolumn{1}{c}{F24} &  60 & 0.70 & 0.75 & \emph{lbest} & 140 & 0.60 & 1.00 & \emph{lbest} & 200 & 0.85 & 0.50 & \emph{lbest}\\ 
\multicolumn{1}{c}{F25} &  25 & 0.90 & 0.50 & \emph{lbest} & 140 & 0.70 & 0.50 & \emph{lbest} & 170 & 0.80 & 0.50 & \emph{lbest} \\ 
\multicolumn{1}{c}{F26} &  80 & 0.55 & 0.75 & \emph{lbest} & 140 & 0.50 & 1.00 & \emph{lbest} & 200 & 0.75 & 1.25 & \emph{lbest} \\ 
\multicolumn{1}{c}{F27} & 100 & 0.85 & 1.25 & \emph{lbest} & 140 & 0.65 & 1.00 & \emph{lbest} & 200 & 0.55 & 1.25 & \emph{lbest}\\ 
\multicolumn{1}{c}{F28} & 100 & 0.85 & 0.75 & \emph{lbest} & 120 & 0.65 & 1.25 & \emph{lbest} & 140 & 0.50 & 1.50 & \emph{lbest} \\ 
\multicolumn{1}{c}{F29} & 100 & 0.55 & 0.50 & \emph{lbest} & 120 & 0.50 & 1.25 & \emph{lbest} & 120 & 0.50 & 0.75 & \emph{lbest}\\ 
\multicolumn{1}{c}{F30} &  40 & 0.50 & 1.75 & \emph{lbest} & 30 & 0.75 & 0.75 & \emph{lbest} & 100 & 0.80 & 0.75 & \emph{lbest}\\ 
\hline
\end{tabular*}
\end{table*}

For comparing the accuracy of the selected PSO variants with the accuracy of their respective sub-variants enabled with LPD, best-of-swarm fitness errors $\varepsilon$ were used, computed with $10^{4}D$ function evaluations (as proposed by the CEC 2014 test \cite{CEC2014}):
\begin{equation}
\label{eq:epsilonmean}
\varepsilon=f_{best}-f^\star~.
\end{equation}
Here $f_{best}$ stands for final best-of-swarm fitness value, averaged across 1000 computational runs, while $f^\star$ stands for known global minimum of goal function $f$. 

Furthermore, so as to provide a relative comparison of the selected methods' performance, a dimensionless rating $\alpha$ is used \cite{LPSO}: 
\begin{equation}
\label{eq:alpha}
\alpha = \frac{\varepsilon_{X}-\varepsilon_{X_L}}{\frac{1}{2}(\varepsilon_{X}+\varepsilon_{X_L})}~,
\end{equation}
where $\varepsilon_{X}$ and $\varepsilon_{X_L}$ represent $\varepsilon$ values for a specific pure PSO variant $X \in \{$  LDIW-PSO, TVAC-PSO, C-PSO, DMS-PSO, CL-PSO$\}$ and its corresponding languid variant $X_L \in \{ $L-LDIW-PSO, L-TVAC-PSO, L-C-PSO, L-DMS-PSO, L-CL-PSO$\}$. A measure of this kind is easy to perceive ($\alpha>0$ means that languid variant performed better than pure variant and vice versa) and may reasonably be averaged across test functions and then used as a bulk value representing overall method score, while its values stay confined to the interval $[-2, 2]$ with $\varepsilon_{X_L}=\varepsilon_{X}$ yielding $\alpha=0$. Note that the definition of \eqref{eq:alpha} is extended so that $\alpha=0$ for $\varepsilon_{X_L}=\varepsilon_{X}$=0.

The results of the prelimary testing of the effects of LPD on selected PSO variants for $D \in \{ 10, 20, 50 \}$ are given in Table \ref{tab:PreliminaryResults}. Here $\alpha_{avg}$ represents the average $\alpha$-value, obtained across all test functions $f \in \{ \textrm{F1}, \textrm{F2}, ..., \textrm{F30} \}$ for a selected PSO variant $X$ and $N_{\alpha^{+}}$ represents the number of test functions with $\alpha>0$, i.e. the number of functions on which the languid variant $X_L$ outperformed the corresponding pure variant $X$.

\begin{table*}[!]\footnotesize
\caption{Preliminary results of testing languid particle dynamics}
\label{tab:PreliminaryResults}
\begin{tabular*}{\textwidth}{l@{\extracolsep{\fill}}rrrrrr}
\hline
\multicolumn{1}{l}{\multirow{2}{*}{Variant}}  & \multicolumn{2}{c}{$D = 10$} & \multicolumn{2}{c}{$D = 20$}  & \multicolumn{2}{c}{$D = 50$} \\
\multicolumn{1}{c}{}  & \multicolumn{1}{c}{$\alpha_{avg}$} & \multicolumn{1}{c}{$N_{\alpha^{+}}$}  & \multicolumn{1}{c}{$\alpha_{avg}$} & \multicolumn{1}{c}{$N_{\alpha^{+}}$} & \multicolumn{1}{c}{$\alpha_{avg}$} & \multicolumn{1}{c}{$N_{\alpha^{+}}$} \\
\hline
LDIW-PSO  &   0.087 &     15  &   0.136 &      17  &      0.143 &     21 \\
TVAC-PSO  &    -0.138 &     8 &  -0.073 &       8  &    -0.109 &     13 \\
C-PSO       &  0.061  &     14   &  0.030  &      17 &     0.189 &     22 \\
DMS-PSO   &   0.096 &     18  &    0.060  &      17 &     0.125 &     19  \\
CL-PSO      &  0.173 &     26   &  0.314 &      29  &     0.382 &      29  \\
\hline
\end{tabular*}
\end{table*}

The results given in Table \ref{tab:PreliminaryResults} allow for several points. First of all, positive values of $\alpha$ for all variants except TVAC-PSO indicate that selected PSO variants, if not PSO in general, benefit from the implementation of LPD; this also corroborates previous research \cite{LPSO}. Moreover, for some of the variants (LDIW-PSO and CL-PSO) the improvement in method accuracy is quite strong. On the other hand, even for the variants whose overall accuracy experienced smaller improvement (C-PSO and DMS-PSO) or even a deterioration (TVAC-PSO), a benefit is still visible for a considerable number of test functions (note in Table \ref{tab:PreliminaryResults} that $N_{\alpha^{+}} \geq 14$, except for TVAC-PSO where $N_{\alpha^{+}} \geq 8$). This means that there exists a rather wide class of optimization problems for which it is reasonable to expect that the employement of languid particle dynamics will produce an improvement in PSO accuracy. 

It may be also interesting to note that the usefulness of LPD mostly increases with the rising of search space dimensionality, indicating that PFIDI techniques may especially be useful for optimization problems with large number of variables.

\subsection{Detailed testing of TVAC-PSO and CL-PSO}
\label{sec:DetailedTesting}

Keeping in mind that PSO parameters should best be understood as problem-related, in order to conduct a comprehensive analysis of the effects of LPD a PSO parameter space exploration procedure was conducted. However, due to substantial computing resources needed for such procedure, only a subselection of the variants used in the preliminary test was included in the detailed testing. In an attempt to `encompass' the preliminary test results (Table \ref{tab:PreliminaryResults}), TVAC-PSO and CL-PSO were chosen because they have shown to gain the least (in the former case) and the most (in the latter case) from introducing LPD.  

In the detailed testing, the chosen variants' performance was tested again on the previously used test functions, although now on a range of PSO parameter combinations. By removing the possibly crucial problem of PSO parameter tuning, it is possible to assess the performance of each variant in a more rigorous manner.
Detailed testing was performed on all combinations of parameters defined by discrete values of swarm size $n$, inertia weight $w_0$ and PSO coefficient $c$ (where $c = c_1 = c_2$), on both \emph{gbest} and \emph{lbest} versions of the PSO method (Table \ref{tab:ParameterRanges}). Note that $c$ is varied only for CL-PSO and PSO version (topology) is only varied for TVAC-PSO.

\begin{table*}[!]\center \footnotesize
	\caption{Method parameters for detailed benchmark testing}
	\label{tab:ParameterRanges}
	\begin{tabular}{p{\dimexpr 0.2\linewidth-2\tabcolsep}>{\centering\arraybackslash}p{\dimexpr 0.8\linewidth-2\tabcolsep}}
		\hline
		\multirow{3}{*}{$n$}
	    & 10,\; 15,\; 20,\; 25,\; 30,\; 40,\; 50,\; 60,\; 80,\; 100 (for $D=10$)\\
		& 20,\; 25,\; 30,\; 40,\; 50,\; 60,\; 80,\; 100,\; 120,\; 140 (for $D=20$) \\
		& 30,\; 40,\; 50,\; 60,\; 80,\; 100,\; 120,\; 140,\; 170,\; 200 (for $D=50$) \\
		\hline
	\end{tabular}
	\begin{tabular}{p{\dimexpr 0.3\linewidth-2\tabcolsep}>{\centering\arraybackslash}p{\dimexpr 0.7\linewidth-2\tabcolsep}}
	$w_0$ & 0.50,\; 0.55,\; 0.60, ..., \; 0.95,\; 1.00 \\
	$c$ {\scriptsize (CL-PSO)} & 0.50,\; 0.75,\; 1.00,\; 1.25,\; 1.50,\; 1.75,\; 2.00 \\
	version {\scriptsize (TVAC-PSO)} & \emph{gbest},\; \emph{lbest} \\ 
	\hline
\end{tabular}
\end{table*}

Given values of method parameters produce a total of 220 and 770 combinations for TVAC-PSO and CL-PSO respectively, all of which were tested on the 30 CEC 2014 test functions, and for the three search space dimensionalities $D \in \{ 10, 20, 50 \}$. For each combination of given parameter values, the optimization result was obtained across 100 computational runs, for both pure TVAC-PSO and CL-PSO variants and the corresponding L-TVAC-PSO and L-CL-PSO variants. (Although CEC 2014 test \cite{CEC2014} proposes using 51 computational runs, authors believe that using 100 computational runs considerably stabilizes the results and makes them significantly more reliable.) On the whole a total of 5.94 million optimization runs per search space dimensionality were executed, making the detailed testing phase of the benchmark testing a considerably large computational effort.

\begin{figure*}
	\includegraphics[width=\linewidth]{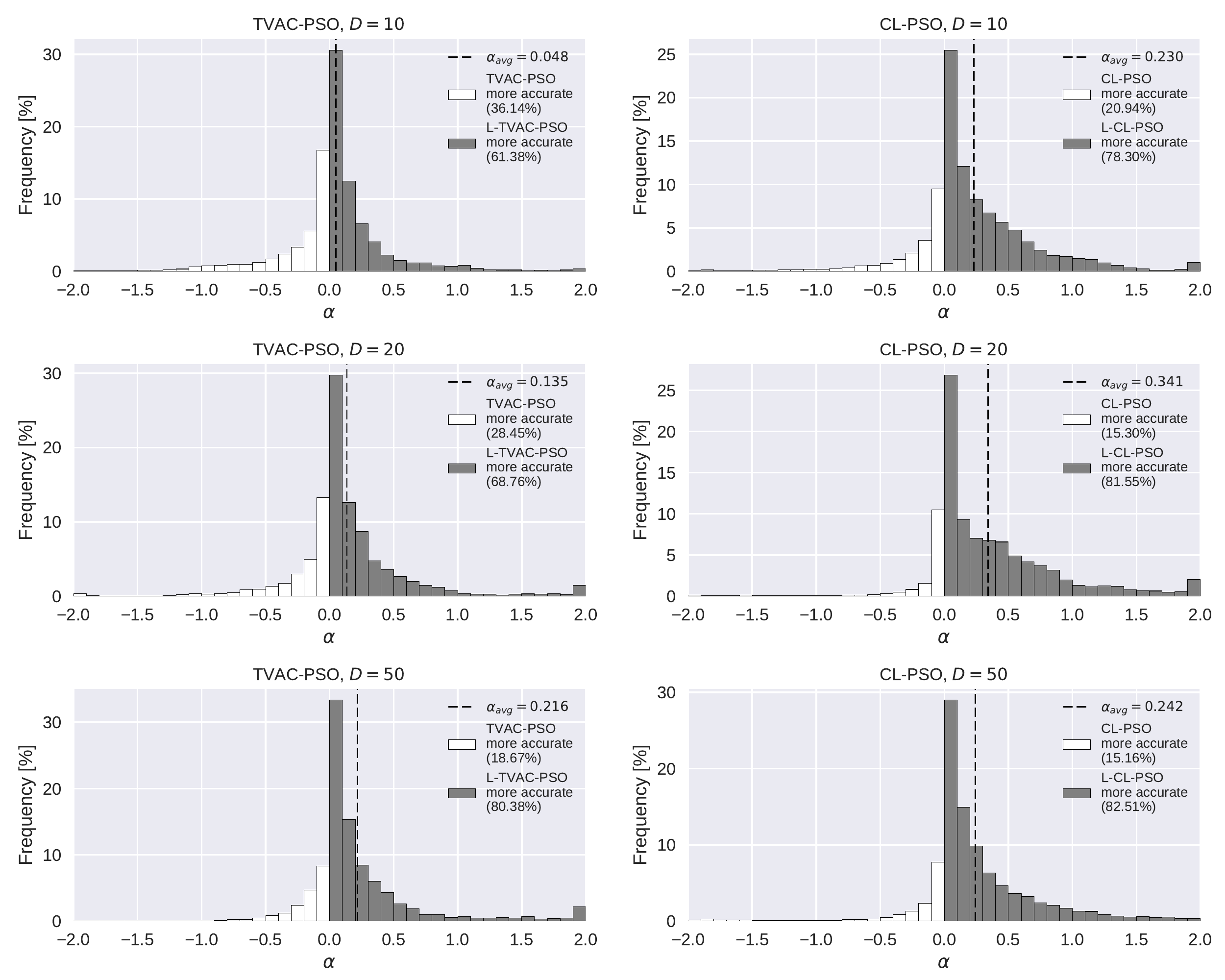}
	\caption{Histograms of final $\alpha$-values for the detailed testing of TVAC-PSO and CL-PSO}
	\label{fig:alphahist}
\end{figure*}

So as to provide some initial statistical overview of the effects of LPD on the accuracy of the tested PSO variants, histograms of final $\alpha$-values (i.e. $\alpha$-values obtained after the last iteration of the optimization process) are given in Figure \ref{fig:alphahist}. These histograms are a representation of the aggregated results of the computations obtained with all used parameter combinations and all test functions.

The histograms in Figure \ref{fig:alphahist} clearly show that implementing LPD in TVAC-PSO and CL-PSO produces measurable improvements in overall method accuracy. As expected on the basis of the preliminary phase of benchmark testing (Section \ref{sec:PreliminaryTesting}), L-CL-PSO yields much stronger improvements over pure CL-PSO than does L-TVAC-PSO over pure TVAC-PSO. Nevertheless, considering the negative preliminary results of TVAC-PSO, it is particulary important that it also demonstrated some improvement in accuracy in the detailed testing. Furthermore, the benefits for accuracy due to LPD consistently increase with rising search space dimensionality, for both variants.

It may be also interesting to note that for some of the method parameter configurations LPD produced dramatic improvements in method accuracy, which is evident by the manifestation of upticks in rightmost bin of the histograms in Figure \ref{fig:alphahist}. Most probably this means that for both PSO variants some specific parameter configurations exist which would normally produce very weak results, and the implementation of LPD serves as a `fix' for some of these situations. Considering that fine-tuning optimization methods in real-world optimization problems is rarely feasible, any modification that reduces the optimization method's sensitivity to its parameters is generally useful.

As a next step in the testing procedure, from the 220 and 770 method parameter combinations used for the detailed benchmark testing of TVAC-PSO and CL-PSO, for each of the used test functions and dimensionalities the best-performing method setups of the pure variant and the corresponding ``languid'' variant were extracted and their results compared. 

In other words, the selected PSO variants and their LPD modifications were fine-tuned in order to compare their performance, eliminating the question of  used method parameters adequacy for each test function or dimensionality.

The comparison of the most accurate TVAC-PSO result against the most accurate L-TVAC-PSO results, for each of the dimenalities $D \in \{10, 20, 50\}$ was given in Tables \ref{tab:TVACBestVsBest10D}, \ref{tab:TVACBestVsBest20D}, and \ref{tab:TVACBestVsBest50D}. Subsequently, the most accurate CL-PSO results was compared with the most accurate L-CL-PSO results in Tables \ref{tab:CLBestVsBest10D}, \ref{tab:CLBestVsBest20D}, and \ref{tab:CLBestVsBest50D}.

Furthermore, statistical analysis of the differences between the best results of variant $X \in \{ $TVAC-PSO, CL-PSO$\}$ and best results of the corresponding ``languid'' variant $X_L \in \{ $L-TVAC-PSO, L-CL-PSO$\}$ was conducted for each test function, on every $D \in \{10, 20, 50\}$. The differences of each of the two corresponding pairs of samples (best $X$ versus best $X_L$) comprising 100 computational runs were checked for statistical significance. If normality of both samples' distribution was confirmed (via Shapiro-Wilk test), the $p$-values were obtained by use of one-sided two-sample t-test, otherwise Wilcoxon rank-sum test was used for test functions with non-normal distributions. In the few instances where $X$ and $X_L$ produced exactly equal results no $p$-value is given.

The obtained $p$-values are also displayed in Tables \ref{tab:TVACBestVsBest10D}-\ref{tab:CLBestVsBest50D}, where  $H_1=X$ stands for the hypothesis $\varepsilon_{X} < \varepsilon_{X_L}$, and $H_1=X_L$ stands for the hypothesis $\varepsilon_{X_L} < \varepsilon_{X}$. The $p$-values printed in boldface belong to the cases for which the difference between the accuracy of the two compared PSO variants was statistically significant ($p<0.05$), i.e. the $H_1$ hypothesis was proven.

\begin{table*}[!]\scriptsize
	\caption{Best TVAC-PSO result versus best L-TVAC-PSO results, $D = 10$}
	\label{tab:TVACBestVsBest10D}
	\begin{tabular*}{\textwidth}{l@{\extracolsep{\fill}}cccccccccc}
		\hline
		\multirow{2}{*}{Fun.}  & \multicolumn{4}{c}{TVAC-PSO ($X$)} & \multicolumn{4}{c}{L-TVAC-PSO ($X_L$)} & \multirow{2}{*}{$H_1$} & \multirow{2}{*}{$p$} \\
		\cline{2-5}\cline{6-9}~ & $n$ & $w_0$ & Variant & $\varepsilon$ & $n$ & $w_0$ & Variant & $\varepsilon$ & ~ & ~ \\ \hline
		F1 & 30 & 0.50 & gbest & $1.3184 \times 10^{3}$ & 20 & 0.75 & gbest & $1.3816 \times 10^{3}$ & $X$ & $0.409^\dagger$\\
		F2 & 100 & 0.50 & lbest & $5.4393 \times 10^{2}$ & 100 & 0.50 & lbest & $5.6891 \times 10^{2}$ & $X$ & $0.410^\dagger$\\
		F3 & 100 & 0.50 & gbest & $3.8649 \times 10^{1}$ & 100 & 0.60 & gbest & $1.4527 \times 10^{2}$ & $X$ & $\mathbf{<0.001}^\dagger$\\
		F4 & 20 & 0.50 & lbest & $2.0846 \times 10^{1}$ & 10 & 0.50 & gbest & $1.9516 \times 10^{1}$ & $X_L$ & $0.199^\dagger$\\
		F5 & 50 & 0.50 & gbest & $1.8291 \times 10^{1}$ & 60 & 0.50 & lbest & $1.7779 \times 10^{1}$ & $X_L$ & $\mathbf{0.025}^\dagger$\\
		F6 & - & - & - & $0.0000 \times 10^{0}$ & - & - & - & $0.0000 \times 10^{0}$ & tie & - \\
		F7 & 100 & 0.50 & lbest & $8.1790 \times 10^{-3}$ & 100 & 0.50 & lbest & $1.0680 \times 10^{-2}$ & $X$ & $\mathbf{0.048}^\dagger$\\
		F8 & 100 & 0.50 & gbest & $1.6019 \times 10^{0}$ & 80 & 0.55 & gbest & $7.5618 \times 10^{-1}$ & $X_L$ & $\mathbf{<0.001}^\dagger$\\
		F9 & 60 & 0.50 & lbest & $4.4649 \times 10^{0}$ & 80 & 0.55 & lbest & $4.0350 \times 10^{0}$ & $X_L$ & $\mathbf{0.021}^\dagger$\\
		F10 & 100 & 0.60 & gbest & $5.7131 \times 10^{1}$ & 50 & 0.60 & gbest & $5.3792 \times 10^{1}$ & $X_L$ & $0.170^\dagger$\\
		F11 & 100 & 0.55 & gbest & $2.4524 \times 10^{2}$ & 100 & 0.50 & gbest & $2.1376 \times 10^{2}$ & $X_L$ & $0.087^\dagger$\\
		F12 & 80 & 0.55 & gbest & $1.3503 \times 10^{-1}$ & 80 & 0.50 & gbest & $1.3396 \times 10^{-1}$ & $X_L$ & $0.441^\dagger$\\
		F13 & 60 & 0.50 & lbest & $4.4080 \times 10^{-2}$ & 100 & 0.50 & lbest & $4.7550 \times 10^{-2}$ & $X$ & $0.175^\dagger$\\
		F14 & 15 & 0.55 & lbest & $9.5050 \times 10^{-2}$ & 40 & 0.55 & lbest & $1.0266 \times 10^{-1}$ & $X$ & $0.136^\dagger$\\
		F15 & 60 & 0.50 & gbest & $7.3838 \times 10^{-1}$ & 50 & 0.50 & gbest & $7.4185 \times 10^{-1}$ & $X$ & $0.382^\dagger$\\
		F16 & 100 & 0.60 & gbest & $1.8872 \times 10^{0}$ & 30 & 0.55 & lbest & $1.7751 \times 10^{0}$ & $X_L$ & $\mathbf{0.045}^\ast$\\
		F17 & 60 & 0.50 & lbest & $1.0864 \times 10^{3}$ & 40 & 0.50 & lbest & $1.1820 \times 10^{3}$ & $X$ & $0.332^\dagger$\\
		F18 & 100 & 0.80 & lbest & $3.7868 \times 10^{2}$ & 100 & 1.00 & lbest & $9.4959 \times 10^{2}$ & $X$ & $\mathbf{<0.001}^\dagger$\\
		F19 & 30 & 0.50 & lbest & $5.2775 \times 10^{-1}$ & 30 & 0.55 & lbest & $6.7133 \times 10^{-1}$ & $X$ & $\mathbf{0.035}^\dagger$\\
		F20 & 100 & 0.70 & gbest & $5.9247 \times 10^{1}$ & 80 & 0.95 & gbest & $1.8969 \times 10^{2}$ & $X$ & $\mathbf{<0.001}^\dagger$\\
		F21 & 20 & 0.60 & lbest & $7.3070 \times 10^{1}$ & 80 & 0.55 & gbest & $7.8336 \times 10^{1}$ & $X$ & $0.385^\dagger$\\
		F22 & 50 & 0.55 & lbest & $1.2958 \times 10^{1}$ & 25 & 0.50 & lbest & $1.3149 \times 10^{1}$ & $X$ & $0.343^\dagger$\\
		F23 & 100 & 0.75 & gbest & $3.2616 \times 10^{2}$ & 100 & 0.90 & lbest & $3.2301 \times 10^{2}$ & $X_L$ & $0.500^\dagger$\\
		F24 & 100 & 0.50 & lbest & $1.0851 \times 10^{2}$ & 100 & 0.50 & lbest & $1.0843 \times 10^{2}$ & $X_L$ & $0.454^\dagger$\\
		F25 & 100 & 0.55 & lbest & $1.2759 \times 10^{2}$ & 100 & 0.60 & lbest & $1.2748 \times 10^{2}$ & $X_L$ & $0.267^\dagger$\\
		F26 & 80 & 0.50 & lbest & $1.0005 \times 10^{2}$ & 100 & 0.50 & lbest & $1.0005 \times 10^{2}$ & $X_L$ & $0.371^\dagger$\\
		F27 & 100 & 0.90 & lbest & $6.8430 \times 10^{0}$ & 100 & 0.70 & lbest & $4.9520 \times 10^{1}$ & $X$ & $0.143^\dagger$\\
		F28 & 100 & 0.50 & lbest & $3.6980 \times 10^{2}$ & 100 & 0.50 & lbest & $3.7067 \times 10^{2}$ & $X$ & $\mathbf{0.038}^\dagger$\\
		F29 & 100 & 0.65 & gbest & $4.2528 \times 10^{2}$ & 80 & 0.55 & gbest & $4.2032 \times 10^{2}$ & $X_L$ & $0.485^\dagger$\\
		F30 & 100 & 0.80 & lbest & $4.7887 \times 10^{2}$ & 100 & 0.90 & lbest & $4.8463 \times 10^{2}$ & $X$ & $\mathbf{0.015}^\dagger$\\\hline
		\multicolumn{11}{l}{$^\ast$ t-test} \\
		\multicolumn{11}{l}{$^\dagger$ Wilcoxon rank-sum test} \\
	\end{tabular*}
\end{table*}

\begin{table*}[!]\scriptsize
	\caption{Best TVAC-PSO result versus best L-TVAC-PSO results, $D = 20$}
	\label{tab:TVACBestVsBest20D}
	\begin{tabular*}{\textwidth}{l@{\extracolsep{\fill}}cccccccccc}
		\hline
		\multirow{2}{*}{Fun.}  & \multicolumn{4}{c}{TVAC-PSO ($X$)} & \multicolumn{4}{c}{L-TVAC-PSO ($X_L$)} & \multirow{2}{*}{$H_1$} & \multirow{2}{*}{$p$} \\
		\cline{2-5}\cline{6-9}~ & $n$ & $w_0$ & Variant & $\varepsilon$ & $n$ & $w_0$ & Variant & $\varepsilon$ & ~ & ~ \\ \hline
		F1 & 20 & 0.50 & lbest & $5.7376 \times 10^{3}$ & 25 & 0.55 & gbest & $6.5384 \times 10^{3}$ & $X$ & $0.322^\dagger$\\
		F2 & - & - & - & $0.0000 \times 10^{0}$ & - & - & - & $0.0000 \times 10^{0}$ & tie & -\\
		F3 & 100 & 0.50 & gbest & $1.9844 \times 10^{1}$ & 100 & 0.55 & gbest & $8.0261 \times 10^{1}$ & $X$ & $\mathbf{<0.001}^\dagger$\\
		F4 & 40 & 0.50 & lbest & $1.2238 \times 10^{-1}$ & 30 & 0.50 & lbest & $4.3322 \times 10^{-1}$ & $X$ & $\mathbf{<0.001}^\dagger$\\
		F5 & 60 & 0.50 & gbest & $2.0006 \times 10^{1}$ & 50 & 0.50 & gbest & $2.0008 \times 10^{1}$ & $X$ & $0.317^\dagger$\\
		F6 & - & - & - & $0.0000 \times 10^{0}$ & - & - & - & $0.0000 \times 10^{0}$ & tie & -\\
		F7 & 140 & 0.55 & lbest & $3.4080 \times 10^{-3}$ & 140 & 0.50 & lbest & $3.5810 \times 10^{-3}$ & $X$ & $0.426^\dagger$\\
		F8 & 120 & 0.55 & gbest & $9.0772 \times 10^{0}$ & 140 & 0.55 & gbest & $6.4076 \times 10^{0}$ & $X_L$ & $\mathbf{<0.001}^\dagger$\\
		F9 & 80 & 0.50 & lbest & $1.5059 \times 10^{1}$ & 140 & 0.60 & gbest & $1.2676 \times 10^{1}$ & $X_L$ & $\mathbf{<0.001}^\dagger$\\
		F10 & 80 & 0.60 & gbest & $2.5089 \times 10^{2}$ & 100 & 0.60 & gbest & $2.0560 \times 10^{2}$ & $X_L$ & $\mathbf{0.005}^\dagger$\\
		F11 & 120 & 0.55 & gbest & $5.0018 \times 10^{2}$ & 120 & 0.50 & gbest & $4.2800 \times 10^{2}$ & $X_L$ & $\mathbf{0.011}^\dagger$\\
		F12 & 80 & 0.50 & gbest & $4.7250 \times 10^{-2}$ & 120 & 0.70 & gbest & $3.5890 \times 10^{-2}$ & $X_L$ & $\mathbf{<0.001}^\dagger$\\
		F13 & 100 & 0.50 & lbest & $1.1818 \times 10^{-1}$ & 120 & 0.50 & lbest & $1.0105 \times 10^{-1}$ & $X_L$ & $\mathbf{<0.001}^\ast$\\
		F14 & 140 & 0.60 & lbest & $1.8996 \times 10^{-1}$ & 120 & 0.55 & lbest & $2.0850 \times 10^{-1}$ & $X$ & $\mathbf{<0.001}^\ast$\\
		F15 & 120 & 0.55 & gbest & $1.9133 \times 10^{0}$ & 120 & 0.50 & gbest & $1.7464 \times 10^{0}$ & $X_L$ & $\mathbf{0.030}^\dagger$\\
		F16 & 60 & 0.55 & gbest & $5.0056 \times 10^{0}$ & 50 & 0.80 & gbest & $4.6811 \times 10^{0}$ & $X_L$ & $\mathbf{0.001}^\ast$\\
		F17 & 20 & 0.50 & gbest & $1.9427 \times 10^{4}$ & 20 & 0.50 & gbest & $2.3047 \times 10^{4}$ & $X$ & $\mathbf{0.043}^\dagger$\\
		F18 & 100 & 0.65 & lbest & $5.1724 \times 10^{3}$ & 140 & 0.60 & lbest & $5.7020 \times 10^{3}$ & $X$ & $\mathbf{0.040}^\dagger$\\
		F19 & 140 & 0.50 & lbest & $2.2182 \times 10^{0}$ & 80 & 0.50 & lbest & $2.3556 \times 10^{0}$ & $X$ & $\mathbf{0.007}^\ast$\\
		F20 & 140 & 0.65 & gbest & $1.7822 \times 10^{2}$ & 100 & 0.80 & gbest & $2.8088 \times 10^{2}$ & $X$ & $\mathbf{<0.001}^\dagger$\\
		F21 & 20 & 0.55 & gbest & $2.5059 \times 10^{3}$ & 20 & 0.50 & gbest & $3.6653 \times 10^{3}$ & $X$ & $\mathbf{<0.001}^\dagger$\\
		F22 & 60 & 0.65 & lbest & $3.4349 \times 10^{1}$ & 50 & 0.70 & lbest & $3.4932 \times 10^{1}$ & $X$ & $0.209^\dagger$\\
		F23 & - & - & - & $3.3006 \times 10^{2}$ & - & - & - & $3.3006 \times 10^{2}$ & tie & -\\
		F24 & 100 & 0.50 & lbest & $2.1018 \times 10^{2}$ & 120 & 0.50 & lbest & $2.1022 \times 10^{2}$ & $X$ & $0.349^\dagger$\\
		F25 & 140 & 0.50 & lbest & $2.0110 \times 10^{2}$ & 120 & 0.95 & lbest & $2.0413 \times 10^{2}$ & $X$ & $\mathbf{<0.001}^\dagger$\\
		F26 & 140 & 0.50 & lbest & $1.0012 \times 10^{2}$ & 140 & 0.50 & lbest & $1.0012 \times 10^{2}$ & $X$ & $0.425^\dagger$\\
		F27 & 120 & 0.55 & lbest & $3.1851 \times 10^{2}$ & 120 & 0.55 & lbest & $3.1639 \times 10^{2}$ & $X_L$ & $0.288^\dagger$\\
		F28 & 100 & 1.00 & lbest & $6.3724 \times 10^{2}$ & 140 & 0.50 & lbest & $6.4794 \times 10^{2}$ & $X$ & $0.159^\dagger$\\
		F29 & 140 & 0.50 & lbest & $3.4475 \times 10^{2}$ & 140 & 0.50 & lbest & $3.5464 \times 10^{2}$ & $X$ & $\mathbf{<0.001}^\dagger$\\
		F30 & 25 & 0.50 & lbest & $9.5986 \times 10^{2}$ & 50 & 0.50 & lbest & $9.0189 \times 10^{2}$ & $X_L$ & $\mathbf{0.015}^\dagger$\\\hline
		\multicolumn{11}{l}{$^\ast$ t-test} \\
		\multicolumn{11}{l}{$^\dagger$ Wilcoxon rank-sum test} \\
	\end{tabular*}
\end{table*}

\begin{table*}[!]\scriptsize
	\caption{Best TVAC-PSO result versus best L-TVAC-PSO results, $D = 50$}
	\label{tab:TVACBestVsBest50D}
	\begin{tabular*}{\textwidth}{l@{\extracolsep{\fill}}cccccccccc}
		\hline
		\multirow{2}{*}{Fun.}  & \multicolumn{4}{c}{TVAC-PSO ($X$)} & \multicolumn{4}{c}{L-TVAC-PSO ($X_L$)} & \multirow{2}{*}{$H_1$} & \multirow{2}{*}{$p$} \\
		\cline{2-5}\cline{6-9}~ & $n$ & $w_0$ & Variant & $\varepsilon$ & $n$ & $w_0$ & Variant & $\varepsilon$ & ~ & ~ \\ \hline
		F1 & 30 & 0.50 & lbest & $5.1583 \times 10^{5}$ & 30 & 0.60 & gbest & $6.1954 \times 10^{5}$ & $X$ & $\mathbf{0.025}^\dagger$\\
		F2 & 170 & 0.55 & lbest & $3.1003 \times 10^{3}$ & 140 & 0.50 & lbest & $2.8774 \times 10^{3}$ & $X_L$ & $0.171^\dagger$\\
		F3 & 100 & 0.65 & gbest & $5.4693 \times 10^{2}$ & 200 & 0.55 & gbest & $1.2044 \times 10^{3}$ & $X$ & $\mathbf{<0.001}^\dagger$\\
		F4 & 50 & 0.55 & lbest & $4.9133 \times 10^{1}$ & 30 & 0.50 & lbest & $6.4531 \times 10^{1}$ & $X$ & $\mathbf{0.005}^\dagger$\\
		F5 & 200 & 0.50 & gbest & $2.0277 \times 10^{1}$ & 140 & 0.50 & gbest & $2.0318 \times 10^{1}$ & $X$ & $0.286^\dagger$\\
		F6 & 170 & 0.55 & lbest & $6.3109 \times 10^{0}$ & 200 & 0.55 & lbest & $4.8182 \times 10^{0}$ & $X_L$ & $\mathbf{<0.001}^\ast$\\
		F7 & 140 & 0.80 & lbest & $3.8570 \times 10^{-3}$ & 100 & 0.85 & lbest & $3.7090 \times 10^{-3}$ & $X_L$ & $0.097^\dagger$\\
		F8 & 200 & 0.60 & gbest & $6.0840 \times 10^{1}$ & 170 & 0.60 & gbest & $3.9158 \times 10^{1}$ & $X_L$ & $\mathbf{<0.001}^\ast$\\
		F9 & 120 & 0.60 & lbest & $9.0608 \times 10^{1}$ & 80 & 0.65 & lbest & $7.1843 \times 10^{1}$ & $X_L$ & $\mathbf{<0.001}^\dagger$\\
		F10 & 200 & 0.55 & gbest & $2.1444 \times 10^{3}$ & 120 & 0.55 & gbest & $1.2998 \times 10^{3}$ & $X_L$ & $\mathbf{<0.001}^\ast$\\
		F11 & 120 & 0.50 & gbest & $5.5401 \times 10^{3}$ & 120 & 0.55 & gbest & $5.2657 \times 10^{3}$ & $X_L$ & $\mathbf{0.008}^\dagger$\\
		F12 & 140 & 0.50 & gbest & $6.6523 \times 10^{-1}$ & 200 & 0.50 & gbest & $5.8612 \times 10^{-1}$ & $X_L$ & $\mathbf{0.014}^\dagger$\\
		F13 & 200 & 0.50 & lbest & $3.2523 \times 10^{-1}$ & 200 & 0.50 & lbest & $2.7795 \times 10^{-1}$ & $X_L$ & $\mathbf{<0.001}^\ast$\\
		F14 & 200 & 0.55 & lbest & $2.5466 \times 10^{-1}$ & 100 & 0.50 & lbest & $2.6835 \times 10^{-1}$ & $X$ & $\mathbf{<0.001}^\dagger$\\
		F15 & 200 & 0.65 & gbest & $1.5266 \times 10^{1}$ & 200 & 0.55 & gbest & $9.4929 \times 10^{0}$ & $X_L$ & $\mathbf{<0.001}^\dagger$\\
		F16 & 80 & 0.50 & gbest & $1.9536 \times 10^{1}$ & 50 & 0.60 & gbest & $1.9478 \times 10^{1}$ & $X_L$ & $0.315^\ast$\\
		F17 & 30 & 0.50 & lbest & $6.3625 \times 10^{4}$ & 30 & 0.50 & lbest & $7.5056 \times 10^{4}$ & $X$ & $\mathbf{0.001}^\dagger$\\
		F18 & 200 & 0.65 & lbest & $5.2746 \times 10^{2}$ & 200 & 0.65 & lbest & $3.0515 \times 10^{2}$ & $X_L$ & $\mathbf{<0.001}^\dagger$\\
		F19 & 120 & 0.65 & gbest & $1.6152 \times 10^{1}$ & 120 & 0.55 & gbest & $1.4257 \times 10^{1}$ & $X_L$ & $\mathbf{<0.001}^\dagger$\\
		F20 & 170 & 0.50 & gbest & $5.6243 \times 10^{2}$ & 140 & 0.70 & gbest & $7.1611 \times 10^{2}$ & $X$ & $\mathbf{<0.001}^\dagger$\\
		F21 & 60 & 0.50 & gbest & $5.1004 \times 10^{4}$ & 50 & 0.50 & gbest & $4.5261 \times 10^{4}$ & $X_L$ & $0.297^\dagger$\\
		F22 & 120 & 0.50 & lbest & $5.1589 \times 10^{2}$ & 170 & 0.50 & lbest & $4.8049 \times 10^{2}$ & $X_L$ & $\mathbf{0.032}^\ast$\\
		F23 & - & - & - & $3.4400 \times 10^{2}$ & - & - & - & $3.4400 \times 10^{2}$ & tie & -\\
		F24 & 200 & 0.50 & lbest & $2.6305 \times 10^{2}$ & 120 & 0.50 & lbest & $2.6406 \times 10^{2}$ & $X$ & $0.143^\dagger$\\
		F25 & 200 & 0.50 & lbest & $2.0204 \times 10^{2}$ & 60 & 0.75 & gbest & $2.0950 \times 10^{2}$ & $X$ & $\mathbf{<0.001}^\dagger$\\
		F26 & 200 & 0.60 & lbest & $1.0030 \times 10^{2}$ & 170 & 0.60 & lbest & $1.0031 \times 10^{2}$ & $X$ & $\mathbf{0.030}^\ast$\\
		F27 & 200 & 0.60 & lbest & $5.0050 \times 10^{2}$ & 200 & 0.55 & lbest & $4.9031 \times 10^{2}$ & $X_L$ & $0.078^\ast$\\
		F28 & 200 & 0.60 & lbest & $1.1603 \times 10^{3}$ & 200 & 0.60 & lbest & $1.1378 \times 10^{3}$ & $X_L$ & $\mathbf{<0.001}^\ast$\\
		F29 & 200 & 0.50 & gbest & $3.1445 \times 10^{6}$ & 140 & 0.50 & gbest & $6.6487 \times 10^{6}$ & $X$ & $\mathbf{0.031}^\dagger$\\
		F30 & 170 & 0.85 & lbest & $8.7158 \times 10^{3}$ & 140 & 0.95 & lbest & $8.6177 \times 10^{3}$ & $X_L$ & $0.226^\dagger$\\\hline
		\multicolumn{11}{l}{$^\ast$ t-test} \\
		\multicolumn{11}{l}{$^\dagger$ Wilcoxon rank-sum test} \\
	\end{tabular*}
\end{table*}

\begin{table*}[!]\scriptsize
	\caption{Best CL-PSO result versus best L-CL-PSO results, $D = 10$}
	\label{tab:CLBestVsBest10D}
	\begin{tabular*}{\textwidth}{l@{\extracolsep{\fill}}cccccccccc}
		\hline
		\multirow{2}{*}{Fun.}  & \multicolumn{4}{c}{CL-PSO ($X$)} & \multicolumn{4}{c}{L-CL-PSO ($X_L$)} & \multirow{2}{*}{$H_1$} & \multirow{2}{*}{$p$} \\
		\cline{2-5}\cline{6-9}~ & $n$ & $w_0$ & $c$ & $\varepsilon$ & $n$ & $w_0$ & $c$ & $\varepsilon$ & ~ & ~ \\ \hline
		F1 & 15 & 0.70 & 0.75 & $6.9497 \times 10^{4}$ & 15 & 0.85 & 1.00 & $5.4072 \times 10^{4}$ & $X_L$ & $\mathbf{0.004}^\dagger$\\
		F2 & 40 & 0.55 & 0.75 & $2.2394 \times 10^{2}$ & 40 & 0.55 & 1.00 & $9.1791 \times 10^{1}$ & $X_L$ & $\mathbf{<0.001}^\dagger$\\
		F3 & 10 & 0.50 & 1.25 & $1.3403 \times 10^{2}$ & 15 & 0.55 & 1.00 & $9.6978 \times 10^{1}$ & $X_L$ & $0.202^\dagger$\\
		F4 & 50 & 0.50 & 0.75 & $4.4994 \times 10^{0}$ & 40 & 0.50 & 1.00 & $1.5506 \times 10^{0}$ & $X_L$ & $\mathbf{0.002}^\dagger$\\
		F5 & 60 & 0.55 & 0.75 & $1.8587 \times 10^{1}$ & 60 & 0.50 & 1.00 & $1.7928 \times 10^{1}$ & $X_L$ & $\mathbf{0.003}^\dagger$\\
		F6 & 20 & 0.50 & 0.75 & $2.6006 \times 10^{-1}$ & 25 & 0.70 & 0.75 & $2.6195 \times 10^{-1}$ & $X$ & $0.265^\dagger$\\
		F7 & 40 & 0.55 & 0.50 & $2.0181 \times 10^{-2}$ & 80 & 0.50 & 0.75 & $8.0740 \times 10^{-3}$ & $X_L$ & $\mathbf{<0.001}^\dagger$\\
		F8 & - & - & - & $0.0000 \times 10^{0}$ & - & - & - & $0.0000 \times 10^{0}$ & tie & -\\
		F9 & 15 & 0.75 & 0.50 & $4.3433 \times 10^{0}$ & 15 & 1.00 & 1.00 & $4.9944 \times 10^{0}$ & $X$ & $\mathbf{0.001}^\ast$\\
		F10 & 15 & 0.55 & 1.00 & $2.5051 \times 10^{1}$ & 20 & 0.55 & 1.00 & $8.6151 \times 10^{0}$ & $X_L$ & $\mathbf{<0.001}^\dagger$\\
		F11 & 10 & 0.50 & 0.75 & $2.5343 \times 10^{2}$ & 10 & 0.75 & 1.00 & $2.2061 \times 10^{2}$ & $X_L$ & $\mathbf{0.049}^\dagger$\\
		F12 & 10 & 0.95 & 2.00 & $4.5657 \times 10^{-1}$ & 10 & 0.55 & 0.50 & $3.9408 \times 10^{-1}$ & $X_L$ & $\mathbf{<0.001}^\ast$\\
		F13 & 25 & 0.70 & 0.50 & $1.1519 \times 10^{-1}$ & 25 & 0.95 & 0.50 & $1.2341 \times 10^{-1}$ & $X$ & $\mathbf{0.012}^\ast$\\
		F14 & 10 & 0.75 & 1.00 & $1.2519 \times 10^{-1}$ & 10 & 0.80 & 1.50 & $1.3755 \times 10^{-1}$ & $X$ & $\mathbf{0.018}^\dagger$\\
		F15 & 10 & 0.75 & 0.50 & $8.5394 \times 10^{-1}$ & 10 & 0.75 & 1.00 & $8.1053 \times 10^{-1}$ & $X_L$ & $0.050^\dagger$\\
		F16 & 10 & 0.75 & 0.50 & $2.0640 \times 10^{0}$ & 15 & 0.75 & 1.00 & $2.0219 \times 10^{0}$ & $X_L$ & $0.223^\ast$\\
		F17 & 15 & 0.70 & 0.75 & $1.8034 \times 10^{3}$ & 20 & 0.90 & 0.75 & $2.3796 \times 10^{3}$ & $X$ & $\mathbf{0.034}^\dagger$\\
		F18 & 50 & 1.00 & 0.50 & $3.0487 \times 10^{2}$ & 60 & 0.70 & 0.50 & $1.7223 \times 10^{2}$ & $X_L$ & $\mathbf{<0.001}^\dagger$\\
		F19 & 15 & 0.55 & 1.00 & $8.8553 \times 10^{-1}$ & 20 & 0.65 & 1.00 & $7.9713 \times 10^{-1}$ & $X_L$ & $\mathbf{0.038}^\dagger$\\
		F20 & 80 & 0.50 & 0.75 & $1.0659 \times 10^{2}$ & 50 & 0.50 & 0.75 & $6.6809 \times 10^{1}$ & $X_L$ & $\mathbf{<0.001}^\dagger$\\
		F21 & 15 & 0.85 & 0.75 & $1.4895 \times 10^{2}$ & 10 & 1.00 & 1.25 & $2.2574 \times 10^{2}$ & $X$ & $\mathbf{0.021}^\dagger$\\
		F22 & 15 & 0.70 & 1.00 & $7.2500 \times 10^{0}$ & 25 & 0.55 & 1.25 & $5.7697 \times 10^{0}$ & $X_L$ & $\mathbf{0.045}^\dagger$\\
		F23 & 100 & 0.55 & 0.75 & $3.1305 \times 10^{2}$ & 100 & 0.50 & 0.75 & $2.9954 \times 10^{2}$ & $X_L$ & $0.083^\dagger$\\
		F24 & 20 & 0.75 & 0.50 & $1.1076 \times 10^{2}$ & 25 & 0.95 & 1.00 & $1.1206 \times 10^{2}$ & $X$ & $\mathbf{<0.001}^\dagger$\\
		F25 & 25 & 0.65 & 1.00 & $1.5628 \times 10^{2}$ & 30 & 0.60 & 1.50 & $1.5053 \times 10^{2}$ & $X_L$ & $\mathbf{0.002}^\dagger$\\
		F26 & 15 & 0.70 & 0.50 & $1.0012 \times 10^{2}$ & 40 & 0.55 & 0.50 & $1.0012 \times 10^{2}$ & $X_L$ & $0.094^\dagger$\\
		F27 & 100 & 0.50 & 0.75 & $3.1503 \times 10^{1}$ & 100 & 0.50 & 0.75 & $3.0256 \times 10^{1}$ & $X_L$ & $\mathbf{0.005}^\dagger$\\
		F28 & 40 & 0.55 & 0.75 & $3.6016 \times 10^{2}$ & 60 & 0.70 & 0.75 & $3.5911 \times 10^{2}$ & $X_L$ & $0.171^\dagger$\\
		F29 & 50 & 0.50 & 0.50 & $4.4976 \times 10^{2}$ & 50 & 0.50 & 0.50 & $4.2645 \times 10^{2}$ & $X_L$ & $\mathbf{0.007}^\dagger$\\
		F30 & 30 & 1.00 & 1.75 & $4.8705 \times 10^{2}$ & 25 & 0.95 & 2.00 & $4.8748 \times 10^{2}$ & $X$ & $0.436^\dagger$\\\hline
		\multicolumn{11}{l}{$^\ast$ t-test} \\
		\multicolumn{11}{l}{$^\dagger$ Wilcoxon rank-sum test} \\
	\end{tabular*}
\end{table*}

\begin{table*}[!]\scriptsize
	\caption{Best CL-PSO result versus best L-CL-PSO results, $D = 20$}
	\label{tab:CLBestVsBest20D}
	\begin{tabular*}{\textwidth}{l@{\extracolsep{\fill}}cccccccccc}
		\hline
		\multirow{2}{*}{Fun.}  & \multicolumn{4}{c}{CL-PSO ($X$)} & \multicolumn{4}{c}{L-CL-PSO ($X_L$)} & \multirow{2}{*}{$H_1$} & \multirow{2}{*}{$p$} \\
		\cline{2-5}\cline{6-9}~ & $n$ & $w_0$ & $c$ & $\varepsilon$ & $n$ & $w_0$ & $c$ & $\varepsilon$ & ~ & ~ \\ \hline
		F1 & 20 & 0.50 & 1.00 & $1.1857 \times 10^{5}$ & 30 & 0.50 & 1.00 & $8.3525 \times 10^{4}$ & $X_L$ & $\mathbf{<0.001}^\dagger$\\
		F2 & - & - & - & $0.0000 \times 10^{0}$ & - & - & - & $0.0000 \times 10^{0}$ & tie & -\\
		F3 & 20 & 0.50 & 1.00 & $1.6850 \times 10^{2}$ & 20 & 0.50 & 1.00 & $1.5163 \times 10^{2}$ & $X_L$ & $0.072^\dagger$\\
		F4 & 20 & 0.50 & 0.75 & $1.1360 \times 10^{1}$ & 25 & 0.55 & 1.00 & $8.1753 \times 10^{0}$ & $X_L$ & $\mathbf{<0.001}^\dagger$\\
		F5 & 20 & 0.90 & 2.00 & $2.0261 \times 10^{1}$ & 20 & 0.65 & 1.00 & $2.0267 \times 10^{1}$ & $X$ & $\mathbf{<0.001}^\dagger$\\
		F6 & 60 & 0.50 & 0.75 & $3.0332 \times 10^{-2}$ & 50 & 0.65 & 0.75 & $7.8269 \times 10^{-2}$ & $X$ & $\mathbf{<0.001}^\dagger$\\
		F7 & 40 & 0.50 & 1.00 & $2.0000 \times 10^{-6}$ & 60 & 0.60 & 1.00 & $0.0000 \times 10^{0}$ & $X_L$ & $0.079^\dagger$\\
		F8 & 20 & 0.60 & 1.00 & $8.6100 \times 10^{-2}$ & 50 & 0.60 & 1.25 & $0.0000 \times 10^{0}$ & $X_L$ & $\mathbf{<0.001}^\dagger$\\
		F9 & 20 & 0.75 & 0.50 & $9.6517 \times 10^{0}$ & 20 & 1.00 & 1.00 & $8.3761 \times 10^{0}$ & $X_L$ & $\mathbf{<0.001}^\dagger$\\
		F10 & 20 & 0.50 & 1.00 & $1.1812 \times 10^{2}$ & 20 & 0.55 & 1.25 & $2.9279 \times 10^{1}$ & $X_L$ & $\mathbf{<0.001}^\dagger$\\
		F11 & 20 & 0.50 & 0.75 & $5.1211 \times 10^{2}$ & 20 & 0.50 & 1.00 & $2.7037 \times 10^{2}$ & $X_L$ & $\mathbf{<0.001}^\dagger$\\
		F12 & 20 & 0.50 & 0.50 & $4.8415 \times 10^{-1}$ & 20 & 0.50 & 0.50 & $3.1218 \times 10^{-1}$ & $X_L$ & $\mathbf{<0.001}^\ast$\\
		F13 & 25 & 0.75 & 0.50 & $1.9500 \times 10^{-1}$ & 50 & 1.00 & 0.75 & $2.0497 \times 10^{-1}$ & $X$ & $\mathbf{0.029}^\ast$\\
		F14 & 20 & 0.75 & 0.50 & $2.8986 \times 10^{-1}$ & 140 & 0.50 & 0.50 & $2.9323 \times 10^{-1}$ & $X$ & $0.283^\ast$\\
		F15 & 20 & 0.75 & 0.50 & $2.7802 \times 10^{0}$ & 25 & 0.80 & 0.75 & $2.4034 \times 10^{0}$ & $X_L$ & $\mathbf{<0.001}^\ast$\\
		F16 & 20 & 0.50 & 0.75 & $5.7072 \times 10^{0}$ & 20 & 0.50 & 1.25 & $4.6711 \times 10^{0}$ & $X_L$ & $\mathbf{<0.001}^\dagger$\\
		F17 & 60 & 0.50 & 0.50 & $8.4474 \times 10^{4}$ & 100 & 0.50 & 0.50 & $7.5716 \times 10^{4}$ & $X_L$ & $0.141^\dagger$\\
		F18 & 20 & 0.65 & 0.75 & $2.7267 \times 10^{3}$ & 30 & 0.50 & 0.75 & $1.0881 \times 10^{3}$ & $X_L$ & $\mathbf{<0.001}^\dagger$\\
		F19 & 20 & 0.70 & 0.50 & $3.0627 \times 10^{0}$ & 20 & 0.95 & 0.50 & $3.2842 \times 10^{0}$ & $X$ & $\mathbf{0.002}^\dagger$\\
		F20 & 20 & 0.65 & 0.75 & $7.8021 \times 10^{2}$ & 25 & 0.60 & 0.75 & $9.3273 \times 10^{2}$ & $X$ & $0.083^\dagger$\\
		F21 & 80 & 0.65 & 0.50 & $1.8743 \times 10^{4}$ & 140 & 0.75 & 0.50 & $1.7422 \times 10^{4}$ & $X_L$ & $0.369^\dagger$\\
		F22 & 20 & 0.85 & 0.50 & $3.4861 \times 10^{1}$ & 50 & 0.90 & 1.00 & $3.6055 \times 10^{1}$ & $X$ & $0.159^\dagger$\\
		F23 & - & - & - & $3.3006 \times 10^{2}$ & - & - & - & $3.3006 \times 10^{2}$ & tie & -\\
		F24 & 20 & 1.00 & 1.25 & $2.0514 \times 10^{2}$ & 40 & 1.00 & 1.50 & $2.0530 \times 10^{2}$ & $X$ & $0.335^\dagger$\\
		F25 & 80 & 0.50 & 0.50 & $2.0113 \times 10^{2}$ & 80 & 0.50 & 0.50 & $2.0068 \times 10^{2}$ & $X_L$ & $\mathbf{0.006}^\dagger$\\
		F26 & 40 & 0.70 & 0.50 & $1.0021 \times 10^{2}$ & 80 & 1.00 & 0.50 & $1.0020 \times 10^{2}$ & $X_L$ & $\mathbf{0.001}^\ast$\\
		F27 & 120 & 0.55 & 0.75 & $3.9394 \times 10^{2}$ & 140 & 0.55 & 0.50 & $3.7433 \times 10^{2}$ & $X_L$ & $\mathbf{<0.001}^\dagger$\\
		F28 & 20 & 0.90 & 1.00 & $6.7885 \times 10^{2}$ & 30 & 1.00 & 1.50 & $6.9379 \times 10^{2}$ & $X$ & $\mathbf{0.008}^\dagger$\\
		F29 & 140 & 0.65 & 1.75 & $3.4313 \times 10^{2}$ & 120 & 0.50 & 1.75 & $3.4865 \times 10^{2}$ & $X$ & $0.351^\dagger$\\
		F30 & 20 & 0.70 & 0.75 & $1.6004 \times 10^{3}$ & 20 & 0.90 & 0.75 & $1.5855 \times 10^{3}$ & $X_L$ & $0.334^\ast$\\\hline
		\multicolumn{11}{l}{$^\ast$ t-test} \\
		\multicolumn{11}{l}{$^\dagger$ Wilcoxon rank-sum test} \\
	\end{tabular*}
\end{table*}

\begin{table*}[!]\scriptsize
	\caption{Best CL-PSO result versus best L-CL-PSO results, $D = 50$}
	\label{tab:CLBestVsBest50D}
	\begin{tabular*}{\textwidth}{l@{\extracolsep{\fill}}cccccccccc}
		\hline
		\multirow{2}{*}{Fun.}  & \multicolumn{4}{c}{CL-PSO ($X$)} & \multicolumn{4}{c}{L-CL-PSO ($X_L$)} & \multirow{2}{*}{$H_1$} & \multirow{2}{*}{$p$} \\
		\cline{2-5}\cline{6-9}~ & $n$ & $w_0$ & $c$ & $\varepsilon$ & $n$ & $w_0$ & $c$ & $\varepsilon$ & ~ & ~ \\ \hline
		F1 & 40 & 0.60 & 0.50 & $7.2154 \times 10^{6}$ & 40 & 0.90 & 0.50 & $5.7225 \times 10^{6}$ & $X_L$ & $\mathbf{<0.001}^\dagger$\\
		F2 & 60 & 0.50 & 0.75 & $1.9987 \times 10^{3}$ & 80 & 0.90 & 0.50 & $1.6386 \times 10^{3}$ & $X_L$ & $0.136^\dagger$\\
		F3 & 30 & 0.50 & 1.00 & $3.5714 \times 10^{3}$ & 30 & 0.50 & 1.25 & $3.2974 \times 10^{3}$ & $X_L$ & $\mathbf{0.020}^\ast$\\
		F4 & 40 & 0.50 & 1.00 & $9.2066 \times 10^{1}$ & 30 & 0.85 & 0.75 & $8.8681 \times 10^{1}$ & $X_L$ & $\mathbf{0.049}^\dagger$\\
		F5 & 30 & 1.00 & 2.00 & $2.0988 \times 10^{1}$ & 30 & 0.50 & 1.00 & $2.1037 \times 10^{1}$ & $X$ & $\mathbf{<0.001}^\dagger$\\
		F6 & 30 & 0.60 & 0.50 & $2.1669 \times 10^{1}$ & 40 & 0.50 & 0.75 & $2.1911 \times 10^{1}$ & $X$ & $0.346^\ast$\\
		F7 & 40 & 0.65 & 1.00 & $1.2000 \times 10^{-5}$ & 60 & 1.00 & 1.25 & $1.6000 \times 10^{-5}$ & $X$ & $0.215^\dagger$\\
		F8 & 30 & 0.80 & 0.50 & $2.6566 \times 10^{1}$ & 30 & 0.50 & 1.25 & $4.5604 \times 10^{0}$ & $X_L$ & $\mathbf{<0.001}^\dagger$\\
		F9 & 30 & 0.75 & 0.50 & $1.3573 \times 10^{2}$ & 30 & 1.00 & 0.75 & $1.5421 \times 10^{2}$ & $X$ & $\mathbf{<0.001}^\ast$\\
		F10 & 30 & 0.50 & 0.75 & $2.5468 \times 10^{3}$ & 30 & 0.50 & 1.00 & $5.8280 \times 10^{2}$ & $X_L$ & $\mathbf{<0.001}^\dagger$\\
		F11 & 30 & 0.50 & 0.50 & $8.8974 \times 10^{3}$ & 30 & 0.50 & 0.50 & $7.9650 \times 10^{3}$ & $X_L$ & $\mathbf{<0.001}^\ast$\\
		F12 & 30 & 1.00 & 2.00 & $1.7195 \times 10^{0}$ & 30 & 0.50 & 0.50 & $1.7767 \times 10^{0}$ & $X$ & $\mathbf{0.042}^\ast$\\
		F13 & 200 & 1.00 & 0.50 & $3.7042 \times 10^{-1}$ & 40 & 1.00 & 0.75 & $3.8033 \times 10^{-1}$ & $X$ & $\mathbf{0.038}^\ast$\\
		F14 & 200 & 0.50 & 0.50 & $3.1848 \times 10^{-1}$ & 170 & 0.60 & 0.50 & $3.0943 \times 10^{-1}$ & $X_L$ & $\mathbf{0.001}^\dagger$\\
		F15 & 30 & 0.75 & 0.50 & $1.9611 \times 10^{1}$ & 30 & 0.95 & 0.75 & $2.0645 \times 10^{1}$ & $X$ & $\mathbf{<0.001}^\ast$\\
		F16 & 30 & 0.65 & 0.50 & $2.1274 \times 10^{1}$ & 30 & 0.55 & 1.00 & $2.0993 \times 10^{1}$ & $X_L$ & $\mathbf{<0.001}^\ast$\\
		F17 & 40 & 0.50 & 0.50 & $8.3558 \times 10^{5}$ & 30 & 0.50 & 0.50 & $6.9178 \times 10^{5}$ & $X_L$ & $\mathbf{0.001}^\dagger$\\
		F18 & 200 & 0.85 & 0.50 & $3.6921 \times 10^{2}$ & 40 & 0.95 & 0.50 & $1.3128 \times 10^{3}$ & $X$ & $\mathbf{<0.001}^\dagger$\\
		F19 & 80 & 0.50 & 1.00 & $2.5370 \times 10^{1}$ & 100 & 0.50 & 1.00 & $2.0383 \times 10^{1}$ & $X_L$ & $\mathbf{<0.001}^\dagger$\\
		F20 & 30 & 0.60 & 0.75 & $2.0757 \times 10^{3}$ & 30 & 0.75 & 0.75 & $2.2256 \times 10^{3}$ & $X$ & $\mathbf{0.042}^\dagger$\\
		F21 & 30 & 0.55 & 0.50 & $7.2866 \times 10^{5}$ & 30 & 0.65 & 0.50 & $5.4988 \times 10^{5}$ & $X_L$ & $\mathbf{<0.001}^\dagger$\\
		F22 & 60 & 0.80 & 0.50 & $4.6012 \times 10^{2}$ & 200 & 0.95 & 0.75 & $5.3035 \times 10^{2}$ & $X$ & $\mathbf{<0.001}^\ast$\\
		F23 & - & - & - & $3.4400 \times 10^{2}$ & - & - & - & $3.4400 \times 10^{2}$ & tie & -\\
		F24 & 40 & 0.50 & 1.75 & $2.5764 \times 10^{2}$ & 30 & 0.55 & 2.00 & $2.5766 \times 10^{2}$ & $X$ & $0.350^\dagger$\\
		F25 & 140 & 0.50 & 0.50 & $2.0919 \times 10^{2}$ & 140 & 0.60 & 0.50 & $2.0191 \times 10^{2}$ & $X_L$ & $\mathbf{<0.001}^\dagger$\\
		F26 & 170 & 0.80 & 0.50 & $1.0042 \times 10^{2}$ & 50 & 1.00 & 1.25 & $1.0053 \times 10^{2}$ & $X$ & $\mathbf{<0.001}^\dagger$\\
		F27 & 80 & 0.55 & 0.50 & $7.8112 \times 10^{2}$ & 80 & 0.50 & 0.75 & $7.7966 \times 10^{2}$ & $X_L$ & $0.436^\dagger$\\
		F28 & 30 & 0.75 & 0.75 & $1.1840 \times 10^{3}$ & 30 & 1.00 & 1.00 & $1.2340 \times 10^{3}$ & $X$ & $\mathbf{<0.001}^\dagger$\\
		F29 & 40 & 0.50 & 1.00 & $4.6058 \times 10^{3}$ & 50 & 0.50 & 1.00 & $4.1448 \times 10^{3}$ & $X_L$ & $\mathbf{0.010}^\dagger$\\
		F30 & 30 & 0.65 & 1.00 & $8.9762 \times 10^{3}$ & 30 & 0.60 & 1.25 & $8.9821 \times 10^{3}$ & $X$ & $0.275^\dagger$\\\hline
		\multicolumn{11}{l}{$^\ast$ t-test} \\
		\multicolumn{11}{l}{$^\dagger$ Wilcoxon rank-sum test} \\
	\end{tabular*}
\end{table*}

The results given in Tables \ref{tab:TVACBestVsBest10D}-\ref{tab:CLBestVsBest50D} may provide some insight into the type of fitness functions which are particularly suitable or not suitable for LPD. With only statistically significant cases taken into account, it may be noted that L-TVAC-PSO seems to be especially successful on test functions F8-F13 and F16. As for L-CL-PSO, it appears to produce best results on a wider range of test functions, namely: F1, F4, F8, F10-F12, F15, F16, F18, F19, F22, F25, F27 and F29. This indicates that LPD may be most favorable for shifted multimodal functions (F4-F16). When implemented in CL-PSO it may also be well-suited for various hybrid or composite types of functions, as well as some simple unimodal functions (e.g. F1, which is an elliptic paraboloid).

\begin{table*}[] \footnotesize
	\centering
	\caption{Summary of the detailed testing results}
	\label{tab:SummaryDetailed}
	\begin{tabular*}{\textwidth}{l@{\extracolsep{\fill}}ccccccc}
		\hline
		\multicolumn{3}{c}{} & \multicolumn{5}{c}{when comparing fine-tuned methods} \\
		\cline{4-8} ~
		Variant & $D$ & $\alpha_{avg}$ & $N_L$ & $\hat{N}_L$ & $N_X$ & $\hat{N}_X$ & $\hat{N}_0$\\
		\hline
		\multirow{3}{*}{TVAC-PSO}
		& 10 & 0.048 & 13 & 4 & 16 & 7 & 19 \\
		& 20 & 0.135 & 10 & 9 & 17 & 10 & 11 \\
		& 50 & 0.216 & 18 & 12 & 11 & 9 & 9 \\
		\hline
		\multirow{3}{*}{CL-PSO}
		& 10 & 0.230 & 21 & 15 & 8 & 6 & 9 \\
		& 20 & 0.341 & 18 & 13 & 10 & 5 & 12 \\
		& 50 & 0.242 & 15 & 13 & 14 & 10 & 7 \\
		\hline
	\end{tabular*}
	\begin{flushleft}
		\noindent\scriptsize
		$N_L$ number of functions where $X_L$ variant performs better than pure $X$ variant\\
		$\hat{N}_L$ number of functions where $X_L$ variant performs significantly better than pure $X$ variant\\
		$N_X$ number of functions where pure $X$ variant performs better than $X_L$ variant\\
		$\hat{N}_X$ number of functions where pure $X$ variant performs significantly better than $X_L$ variant\\
		$\hat{N}_0$ number of functions with no significant difference in performance of $X_L$ and pure $X$ variant
	\end{flushleft}
\end{table*}

A summary of the results of the detailed testing of TVAC-PSO versus L-TVAC-PSO and CL-PSO versus L-CL-PSO is given in Table \ref{tab:SummaryDetailed}, which allows for several observations. Firstly, although $\alpha_{avg}$ does mostly increase with $D$, when comparing fine-tuned PSO variants there are no clear indications that the advantages of languid variants remain equally strong after fine-tuning the method parameters. However, the implementation of LPD provides significant improvements in method accuracy for up to 50\% of the CEC 2014 test functions, showing that there is a wide class of optimization problems which are well-suited for languid PSO methods. Moreover, LPD significantly deteriorated TVAC-PSO and CL-PSO accuracy for never more than 33\% of the test functions, making LPD a technique with low liability for possible impairment of PSO performance.

\section{Conclusion}
\label{sec:Conclusion}

As a continuation of previous research, the effects of enhanced particle `self-awareness' in terms of Personal Fitness Improvement Dependent Inertia (PFIDI) on the accuracy of a selection of PSO variants were explored. The PFIDI method used for this  was languid particle dynamics (LPD), which makes inertia a conditional term in PSO velocity update, enabled only for particles which managed to improve their position in the previous iteration.

Five PSO variants were selected for testing the effects of LPD: standard PSO with linearly decreasing inertia weight (LDIW-PSO), time varying acceleration coefficients PSO (TVAC-PSO), Chaotic PSO (C-PSO), Dynamic multiswarm PSO (DM-PSO) and Comprehensive learning PSO (CL-PSO). Each of the selected variants was tested in preliminary benchmark testing on 30 test functions (CEC 2014) and three dimensionalities $D \in \{ 10, 20, 50 \}$ by comparing their accuracy with the accuracy of the corresponding languid variant, where all but one have responded positively. The best and worst performing variants (namely, CL-PSO and TVAC-PSO) were additionally scrutinized in the detailed testing phase, which included testing across a spectrum of method parameter configurations, as well as the comparison of fine-tuned methods and their languid counterparts. Finally, the obtained results were statistically validated against the significance level of 0.05.

The results have shown that both TVAC-PSO and CL-PSO, and by extension probably all five selected PSO variants, gain visible advances in accuracy when enhanced with LPD. The tested languid PSO variants produce better accuracy over the corresponding pure PSO variants in 33-70\% of test functions, with statistically significant improvements in accuracy demonstrated for 13-50\% of test functions. With the rising of problem  dimensionality, the benefits on accuracy mostly rise as well, indicating that PFIDI techniques may be particularly useful for high-dimension problems. The testing procedure included 30 test functions, of which multimodal functions of relatively modest complexity have shown to be most suitable for LPD.

The implementation of languid particle dynamics also somewhat stabilizes the method performance with regards to method parameters, making it more robust and reliable, which is especially useful in real-world optimization. Furthermore, considering that at any time not more a third of test functions have shown to yield poorer performance after implementation of LPD means that it would be possibly reasonable for many PSO variants to have LPD implemented by default, at least when used in general-purpose optimization codes.

Potentially even more interesting lines of research on PFIDI still remain to be explored. These include detecting more sophisticated and/or more efficient methods of particle-wise fitness-based inertia handling, as well as investigating the possibilities of generalizing PFIDI in a way which would allow for continuous dynamic inertia adaptation based on personal fitness improvement.



%
%

\bibliography{bibliography}

\begin{thebibliography}{10}
\expandafter\ifx\csname url\endcsname\relax
  \def\url#1{\texttt{#1}}\fi
\expandafter\ifx\csname urlprefix\endcsname\relax\def\urlprefix{URL }\fi
\expandafter\ifx\csname href\endcsname\relax
  \def\href#1#2{#2} \def\path#1{#1}\fi

\bibitem{PSO1}
J.~Kennedy, R.~Eberhart, Particle swarm optimisation. 1995, in: Proceedings
  IEEE International Conference on Neural Networks, Vol.~IV, IEEE Service
  Center, Piscataway, NJ, 1995, pp. 1942--1948.

\bibitem{PSO2}
R.~Eberhart, J.~Kennedy, A new optimizer using particle swarm theory, in: Micro
  Machine and Human Science, 1995. MHS'95., Proceedings of the Sixth
  International Symposium on, IEEE, 1995, pp. 39--43.

\bibitem{LPSO}
S.~Dru{\v{z}}eta, S.~Ivi{\'{c}},
  \href{http://dx.doi.org/10.1007/s00500-015-2016-7}{Examination of benefits of
  personal fitness improvement dependent inertia for particle swarm
  optimization}, Soft Computing (2016) 1--14\href
  {https://doi.org/10.1007/s00500-015-2016-7}
  {\path{doi:10.1007/s00500-015-2016-7}}.
\newline\urlprefix\url{http://dx.doi.org/10.1007/s00500-015-2016-7}

\bibitem{BRAENDLER}
D.~Braendler, T.~Hendtlass,
  \href{http://dx.doi.org/10.1007/3-540-48035-8\_19}{The suitability of
  particle swarm optimisation for training neural hardware}, in: T.~Hendtlass,
  M.~Ali (Eds.), Developments in Applied Artificial Intelligence, Vol. 2358 of
  Lecture Notes in Computer Science, Springer Berlin Heidelberg, 2002, pp.
  190--199.
\newblock \href {https://doi.org/10.1007/3-540-48035-8\_19}
  {\path{doi:10.1007/3-540-48035-8\_19}}.
\newline\urlprefix\url{http://dx.doi.org/10.1007/3-540-48035-8\_19}

\bibitem{FDRPSO}
K.~Veeramachaneni, T.~Peram, C.~Mohan, L.~A. Osadciw,
  \href{http://dl.acm.org/citation.cfm?id=1761233.1761244}{Optimization using
  particle swarms with near neighbor interactions}, in: Proceedings of the 2003
  international conference on Genetic and evolutionary computation: PartI,
  GECCO'03, Springer-Verlag, Berlin, Heidelberg, 2003, pp. 110--121.
\newline\urlprefix\url{http://dl.acm.org/citation.cfm?id=1761233.1761244}

\bibitem{IWSTRAT}
J.~Bansal, P.~Singh, M.~Saraswat, A.~Verma, S.~S. Jadon, A.~Abraham, Inertia
  weight strategies in particle swarm optimization, in: Nature and Biologically
  Inspired Computing (NaBIC), 2011 Third World Congress on, IEEE, 2011, pp.
  633--640.

\bibitem{CLERC}
M.~Clerc, \href{http://clerc.maurice.free.fr/pso}{Think locally, act locally:
  The way of life of cheap-pso, an adaptive pso}, Tech. rep., Technical Report
  (2001).
\newline\urlprefix\url{http://clerc.maurice.free.fr/pso}

\bibitem{CPSO}
B.~Liu, L.~Wang, Y.-H. Jin, F.~Tang, D.-X. Huang, Improved particle swarm
  optimization combined with chaos, Chaos, Solitons \& Fractals 25~(5) (2005)
  1261--1271.

\bibitem{RAGHAVENDRA}
R.~Raghavendra, B.~Dorizzi, A novel adaptive inertia particle swarm
  optimization (aipso) algorithm for improving multimodal biometric
  recognition, in: Hand-Based Biometrics (ICHB), 2011 International Conference
  on, 2011, pp. 1--6.
\newblock \href {https://doi.org/10.1109/ICHB.2011.6094299}
  {\path{doi:10.1109/ICHB.2011.6094299}}.

\bibitem{SCORE}
X.~Cai, Z.~Cui, J.~Zeng, Y.~Tan, Performance-dependent adaptive particle swarm
  optimization, International Journal of Innovative Computing Information and
  Control 3~(6 B) (2007) 1697--1706.

\bibitem{DONG}
C.~Dong, G.~Wang, Z.~Chen, Z.~Yu, A method of self-adaptive inertia weight for
  pso, in: Computer Science and Software Engineering, 2008 International
  Conference on, Vol.~1, 2008, pp. 1195--1198.
\newblock \href {https://doi.org/10.1109/CSSE.2008.295}
  {\path{doi:10.1109/CSSE.2008.295}}.

\bibitem{IAPSO}
K.~Suresh, S.~Ghosh, D.~Kundu, A.~Sen, S.~Das, A.~Abraham, Inertia-adaptive
  particle swarm optimizer for improved global search, in: Intelligent Systems
  Design and Applications, 2008. ISDA'08. Eighth International Conference on,
  Vol.~2, IEEE, 2008, pp. 253--258.

\bibitem{FENG}
C.~Feng, S.~Cong, X.~Feng, A new adaptive inertia weight strategy in particle
  swarm optimization, in: Evolutionary Computation, 2007. CEC 2007. IEEE
  Congress on, IEEE, 2007, pp. 4186--4190.

\bibitem{ZHANG}
X.~Zhang, Y.~Du, G.~Qin, Adaptive particle swarm algorithm with dynamically
  changing inertia weight, J. Xian Jiaotong Univ. 39 (2005) 1039--1042.

\bibitem{YANG}
X.~Yang, J.~Yuan, J.~Yuan, H.~Mao, A modified particle swarm optimizer with
  dynamic adaptation., Applied Mathematics and Computation 189~(2) (2007)
  1205--1213.

\bibitem{DYNAMIC}
I.~Rezazadeh, M.~R. Meybodi, A.~Naebi, Adaptive particle swarm optimization
  algorithm for dynamic environments, in: Advances in swarm intelligence,
  Springer, 2011, pp. 120--129.

\bibitem{LAIW}
K.~Deep, M.~Arya, J.~C. Bansal,
  \href{http://doi.acm.org/10.1145/2001576.2001732}{A non-deterministic
  adaptive inertia weight in pso}, in: Proceedings of the 13th annual
  conference on Genetic and evolutionary computation, GECCO '11, ACM, New York,
  NY, USA, 2011, pp. 1155--1162.
\newblock \href {https://doi.org/10.1145/2001576.2001732}
  {\path{doi:10.1145/2001576.2001732}}.
\newline\urlprefix\url{http://doi.acm.org/10.1145/2001576.2001732}

\bibitem{PSO98}
Y.~Shi, R.~Eberhart, A modified particle swarm optimizer, in: Evolutionary
  Computation Proceedings, 1998. IEEE World Congress on Computational
  Intelligence., The 1998 IEEE International Conference on, IEEE, 1998, pp.
  69--73.

\bibitem{PHDTHESIS}
F.~van~den Bergh, An analysis of particle swarm optimizers., submitted ph. d,
  Ph.D. thesis, thesis, University of Pretoria, Pretoria (2001).

\bibitem{STANDARDPSO}
D.~Bratton, J.~Kennedy, Defining a standard for particle swarm optimization,
  in: Swarm Intelligence Symposium, 2007. SIS 2007. IEEE, IEEE, 2007, pp.
  120--127.

\bibitem{TVAC}
A.~Ratnaweera, S.~K. Halgamuge, H.~C. Watson, Self-organizing hierarchical
  particle swarm optimizer with time-varying acceleration coefficients, IEEE
  Transactions on evolutionary computation 8~(3) (2004) 240--255.

\bibitem{PSOCODE}
PSC, \href{http://www.particleswarm.info/Standard\_PSO\_2006.c}{Particle swarm
  central, standard {PSO} 2006} (2006).
\newline\urlprefix\url{http://www.particleswarm.info/Standard\_PSO\_2006.c}

\bibitem{LDIW}
Y.~Shi, R.~C. Eberhart, Empirical study of particle swarm optimization, in:
  Proceedings of the 1999 Congress on Evolutionary Computation-CEC99 (Cat. No.
  99TH8406), Vol.~3, 1999, p. 1950 Vol. 3.
\newblock \href {https://doi.org/10.1109/CEC.1999.785511}
  {\path{doi:10.1109/CEC.1999.785511}}.

\bibitem{DMSPSO}
J.-J. Liang, P.~N. Suganthan, Dynamic multi-swarm particle swarm optimizer with
  local search, in: 2005 IEEE Congress on Evolutionary Computation, Vol.~1,
  Ieee, 2005, pp. 522--528.

\bibitem{CLPSO}
J.~J. Liang, A.~K. Qin, P.~N. Suganthan, S.~Baskar, Comprehensive learning
  particle swarm optimizer for global optimization of multimodal functions,
  IEEE transactions on evolutionary computation 10~(3) (2006) 281--295.

\bibitem{CEC2014}
J.~Liang, B.~Qu, P.~Suganthan, Problem definitions and evaluation criteria for
  the {CEC} 2014 special session and competition on single objective
  real-parameter numerical optimization, Computational Intelligence Laboratory,
  Zhengzhou University, Zhengzhou China and Technical Report, Nanyang
  Technological University, Singapore.

\bibitem{SPSO}
M.~Clerc,
  \href{http://clerc.maurice.free.fr/pso/SPSO\_descriptions.pdf}{Standard
  particle swarm optimisation from 2006 to 2011} (2012).
\newline\urlprefix\url{http://clerc.maurice.free.fr/pso/SPSO\_descriptions.pdf}

\end{thebibliography}

\end{document}